\documentclass{article}
\usepackage[final]{colm2026_conference}

\usepackage[utf8]{inputenc}
\usepackage[T1]{fontenc}
\usepackage{xcolor}
\usepackage{microtype}
\usepackage{url}
\usepackage{booktabs}
\usepackage{amsfonts}
\usepackage{nicefrac}
\usepackage{graphicx}
\usepackage{subfigure}
\usepackage{caption}
\usepackage{pifont}
\usepackage{amsmath}
\usepackage{amssymb}
\usepackage{mathtools}
\usepackage{amsthm}
\usepackage{algorithm}
\usepackage{algorithmic}
\usepackage{paperpkg}
\usepackage{wrapfig}
\usepackage{etoc}
\usepackage[title]{appendix}
\usepackage{makecell}
\usepackage{nicematrix}
\usepackage{lineno}
\usepackage[colorlinks]{hyperref}
\usepackage[capitalize,noabbrev]{cleveref}

\definecolor{darkblue}{rgb}{0, 0, 0.5}
\definecolor{tablebg}{RGB}{220,230,245}
\hypersetup{colorlinks=true, citecolor=[HTML]{3366CC}, linkcolor=darkblue, urlcolor=darkblue}

\newcommand{\alg}{{{NIRVANA}}}

\newcommand{\qscore}[4]{%
  {\scriptsize F:#1, R:#2, C:#3, H:#4}%
}

\theoremstyle{plain}
\newtheorem{theorem}{Theorem}[section]
\newtheorem{proposition}[theorem]{Proposition}

\theoremstyle{definition}

\theoremstyle{remark}

\author{Mengting Ai, Tianxin Wei, Sirui Chen, Jingrui He \\
University of Illinois Urbana-Champaign\\
\texttt{\{mai10, jingrui\}@illinois.edu} 
}

\title{NIRVANA: Structured Pruning Reimagined for Large Language Model Compression}

\begin{document}

\ifcolmsubmission
\linenumbers
\fi

\maketitle

\begin{abstract}
While structured pruning presents a highly effective pathway for accelerating Large Language Model (LLM) inference, existing methods frequently suffer from significant performance degradation and demand computationally retraining to recover capabilities. To overcome these barriers, we present \alg, a novel, hardware-aware structured pruning framework designed to preserve both zero-shot performance and the optimization landscape for downstream fine-tuning. Departing from traditional loss-based heuristics, our approach evaluates structural importance through a first-order function-space saliency inspired by the Neural Tangent Kernel (NTK), effectively safeguarding the model's critical training dynamics. To prevent structural collapse at high compression rates, we introduce a global unit-ranking strategy coupled with an analytically derived allocation mechanism, which optimally balances the pruning aggressiveness between attention heads and MLP neurons. Furthermore, we eliminate the instability typically associated with random data sampling by employing a lightweight, KL-divergence-driven calibration data selection process. Extensive evaluations across Llama3, Qwen, and T5 architectures demonstrate that \alg~consistently establishes new state-of-the-art results on different benchmarks, providing a theoretically sound and practical approach to LLM compression. 
\end{abstract}

\addtocontents{toc}{\protect\setcounter{tocdepth}{-1}}

\section{Introduction}

While Transformer-based large language models (LLMs) \citep{vaswani2017attention} have revolutionized natural language processing, their enormous computational requirements pose a significant barrier to widespread adoption. For example, a standard 7B-parameter model requires $\sim$14GB of GPU memory at 16-bit precision, incurring prohibitive costs that often restrict these advanced AI tools to resource-rich institutions \citep{ai2025mlpfusionefficientfinetuning,zou2024promptinternsavinginferencecosts,dang2025fzoofastzerothorderoptimizer,pmlr-v202-wei23b,10.1145/3701551.3703577}.

To alleviate this critical bottleneck, model compression techniques, particularly pruning \citep{lecun1989generalization}, emerge as an essential strategy, aiming to create lighter, more accessible models without substantially compromising their effectiveness. Current pruning approaches generally fall into three categories:
\textbf{(1) Unstructured pruning} methods (e.g., SparseGPT~\citep{pmlr-v202-frantar23a}, Wanda~\citep{sun2023wanda}\footnote{These two can also be applied for semi-structured pruning.})
achieve near-lossless zero-shot accuracy by removing individual weights \citep{pmlr-v202-frantar23a,sun2023wanda}, but fail to deliver practical speedups due to irregular sparsity patterns incompatible with hardware accelerators \citep{10643325}.
\textbf{(2) Semi-structured pruning} (e.g., 2:4 block sparsity~\citep{zheng2024learnefficientbuildstructured}) addresses this limitation by enforcing fixed sparsity patterns optimized for NVIDIA sparse tensor cores \citep{Yang_2023_CVPR}.
However, such approaches still struggle during supervised fine-tuning (SFT), as optimizer updates inevitably disrupt the predefined structures, limiting their end-to-end usability.
\textbf{(3) Structured pruning} methods, such as LLM-Pruner~\citep{NEURIPS2023_44956951} and FLAP~\citep{flap},
further improve hardware compatibility by removing entire neurons or layers, offering acceleration across both inference and training.
Orthogonal to pruning, quantization further compresses models by reducing their numerical precision, presenting a similar dilemma between computational gains and performance preservation \citep{MLSYS2024_42a452cb,frantar2023gptq,pmlr-v202-xiao23c,zhao2024sparsevqtransformerffnfreeframework}.

While structured pruning holds the greatest promise for practical deployment, existing methods face several critical challenges:
\textbf{(1) Inefficient recovery tuning lacking alignment with fine-tuning dynamics:} 
To recover from pruning-induced performance drops, existing methods rely heavily on costly fine-tuning procedures like LoRA adapters~\citep{hu2021loralowrankadaptationlarge} or extensive SFT~\citep{xia2024shearedllamaacceleratinglanguage}. However, these recovery methods do not explicitly account for how pruning decisions influence the model's fine-tuning capability, resulting in inefficient resource usage and sub-optimal results.
\textbf{(2) Ignoring layer- and module-specific characteristics:} Current methods typically apply pruning uniformly across all layers or modules (attention and MLP), disregarding their distinct roles within the network.
This oversight often results in suboptimal pruning choices, impairing model performance.
\textbf{(3) Neglect of calibration data influence:} Pruning decisions depend critically on the calibration dataset used; yet, existing approaches rarely discuss or optimize this crucial factor, leaving pruning outcomes vulnerable to suboptimal data selection.

To address these critical gaps, we introduce \textbf{NIRVANA}
(\textbf{N}TK-\textbf{I}nfo\textbf{R}med adapti\textbf{V}e neuron \& \textbf{A}ttentio\textbf{N} he\textbf{A}d pruning), 
a novel structured pruning method that tightly integrates pruning decisions with model fine-tuning dynamics through the lens of the Neural Tangent Kernel (NTK)~\citep{jacot2018neural}. By aligning pruning criteria with the NTK spectrum under Adam: the de facto optimizer for LLMs, NIRVANA uniquely balances immediate accuracy preservation and long-term fine-tuning adaptability. Additionally, NIRVANA employs an adaptive sparsity allocation strategy across layers and modules, complemented by a calibration data selection mechanism based on KL divergence, making pruning both theoretically grounded and practically effective. Our primary contributions are:
\begin{itemize}[itemsep=0pt, parsep=1pt, topsep=0pt,leftmargin=2em]
\item \textbf{NIRVANA}: A novel NTK-guided structured pruning method
explicitly designed to preserve zero-shot accuracy while maintaining fine-tuning capability, connecting pruning decisions to training dynamics.
\item An \textbf{adaptive sparsity allocation} strategy that dynamically adjusts pruning ratios across layers and modules, explicitly addressing overlooked disparities in existing pruning methodologies.
\item A \textbf{KL-divergence-driven calibration data selection} strategy ensuring pruning robustness by identifying optimal subsets that minimize post-pruning output discrepancies, effectively decoupling pruning quality from calibration dataset size.
\item \textbf{Comprehensive experiments} on prominent LLMs (Llama3 family, Qwen and T5) 
demonstrating that NIRVANA significantly outperforms state-of-the-art structured pruning methods in perplexity and downstream task accuracy under similar sparsity budgets, while seamlessly integrating into standard fine-tuning pipelines without requiring additional modifications.
\end{itemize}

\section{Preliminary \& Problem Formulation }

\textbf{Preliminary of Model Architecture and NTK.}
We use lowercase letters to denote scalars, boldface lowercase letters to denote vectors, and boldface uppercase letters to denote matrices. The element-wise product is denoted by \(\odot\). The neural network is denoted by \(f\), parameterized by \(\mathbf{W}\), and \(\mathbf{x}\) represents the input data. See the full notation in \cref{tab:notation}.
Since most of the current LLMs are based on SwiGLU \citep{shazeer2020gluvariantsimprovetransformer} structure, we focus on pruning the Attention \& MLP sub-layer inside a SwiGLU Transformer block. For the MLP block, it is parameterized by three weight matrices: $\mtx{W}_\text{gate} \in \mathbb{R}^{d \times m} \, ,\mtx{W}_\text{up} \in \mathbb{R}^{d \times m} \, ,\text{and} \, \mathbf{W}_\text{down} \in \mathbb{R}^{m \times d}.$
For an input token \(\mathbf{x} \in \mathbb{R}^{d}\), the MLP output is computed as:
\begin{equation*}
\mathbf{H}(\mathbf{x})=\Bigl(\sigma(\mathbf{x}\mathbf{W}_\text{gate}) \odot \mtx x\mtx{W}_\text{up} \Bigr)\mathbf{W}_\text{down},
\end{equation*}
where the activation function Swish \(\sigma(\cdot)\) is applied elementwise, and biases are omitted for simplicity (either because modern designs often exclude them or their contribution is marginal \citep{dubey2024llama3herdmodels}). As for the Multi-Head Attention (MHA) block with 
$\mtx{Q}_a = \mtx{x} \mtx{W}_a^{Q}, \mtx{K}_a = \mtx{x} \mtx W_a^{K}, \mtx V_a = \mtx{x} \mtx W_a^{V}, a=1,\dots,h$, 
where $\mtx W_a^{Q}, \mtx W_a^{K}, \mtx W_a^{V} \in \mathbb{R}^{d \times d_h}$ and $d_h = d/h$, the attention for each head $a$ is $\text{head}_a = \text{softmax}\left(\frac{\mtx Q_a \mtx K_a^\top}{\sqrt{d_h}}\right) \mtx V_a,$
and the MHA output \footnote{In Llama3's implementation, which employs Grouped Query Attention (GQA), multiple query heads share the same key-value (KV) head. When calculating group saliency scores, we align query (Q) and output (O) heads with their corresponding KV head groupings through index mapping.} is: 
\begin{equation*}    
\text{MHA}(\mtx x) = [\text{head}_1, \dots, \text{head}_h] \mtx W^O, \quad \mtx W^O \in \mathbb{R}^{h d_h \times d}.
\end{equation*}
Neural Tangent Kernel (NTK) \citep{jacot2018neural} provides a kernel-based framework for analyzing the training dynamics of neural networks by approximating their behavior as linear models in function space.
In our paper, we use Adam-based NTK in the form of: 
\begin{align*}
{\Theta}(\mtx x,\mtx x)
&=
\nabla_{\mathbf{W}}f(\mtx x;\mathbf{W})\,
\mathrm{sign}\bigl(\nabla_{\mathbf{W}}f(\mtx x;\mathbf{W})\bigr)^\top
=
\bigl\langle
  \nabla_{\mathbf{W}}f(\mtx x;\mathbf{W}),
  \mathrm{sign}\bigl(\nabla_{\mathbf{W}}f(\mtx x;\mathbf{W})\bigr)
\bigr\rangle.
\end{align*}
See the details of the derivation in \cref{app:ntk}

\textbf{Problem Formulation of Pruning Process.}
\label{sec:problem_formulation}
We assume that the output \(f(\mathbf{x};\mathbf{W})\) is a single value, which is common in classification or next-token generation tasks.\footnote{Without loss of generality, our analysis can be extended to the vector-output case.} Given a sparsity level $v$, the target of pruning can be then written in the form of:
\begin{align*}
\text{argmin}_{\hat{\mathbf{W}},\mathbf{M}}\mathcal{L}\Bigl(f(\mathbf{x};\mathbf{W}),\,f(\mathbf{x};\hat{\mathbf{W}}\odot\mathbf{M})\Bigr),
\end{align*}
where $\mtx M$ 
is the mask matrix that has the same shape as $\mtx W$.
Directly solving this joint optimization over \(\hat{\mathbf{W}}\) and \(\mathbf{M}\) is NP-hard. Consequently, popular practices include fixing the weights (i.e., setting \(\hat{\mathbf{W}}=\mathbf{W}\)) and searching for \(\mathbf{M}\) only (one-shot pruning \citep{frankle2018the,pmlr-v202-frantar23a,sun2023wanda,NEURIPS2021_a376033f}), or selecting \(\mathbf{M}\) first and then optimizing \(\hat{\mathbf{W}}\) , which typically requires further fine-tuning or re-training \citep{NEURIPS2022_987bed99,NEURIPS2023_44956951}. General pruning approaches define a \emph{saliency score} \(S_{i,j}\) for each weight, which estimates the impact of removing the connection \(\mathbf{W}_{i,j}\). A general form of the saliency score is:
\begin{equation}
\label{eq:general_saliency}
S_{i,j}
= \frac{\partial \mathcal{I}}{\partial \mathbf{W}_{i,j}} \cdot \mathbf{W}_{i,j},
\end{equation}
where \(\mathcal{I}\) is a function that measures the importance of the weight \(\mathbf{W}\) to the network \(f\). 
Once these scores are computed, the mask is obtained by selecting the top \(\kappa\%\) of weights:
\[
\mathbf{M}_{i,j} = \text{Top}_{\kappa}(S)_{i,j} = \begin{cases}
1, & \text{if } S_{i,j} \text{ is among the top } \kappa\%,\\[1mm]
0, & \text{otherwise}.
\end{cases}
\]
This binary mask is then applied to the weights for pruning.

\begin{figure*}[t]
  \centering
  \includegraphics[width=0.9\linewidth, trim=0.02cm 0 0.02cm 0, clip]{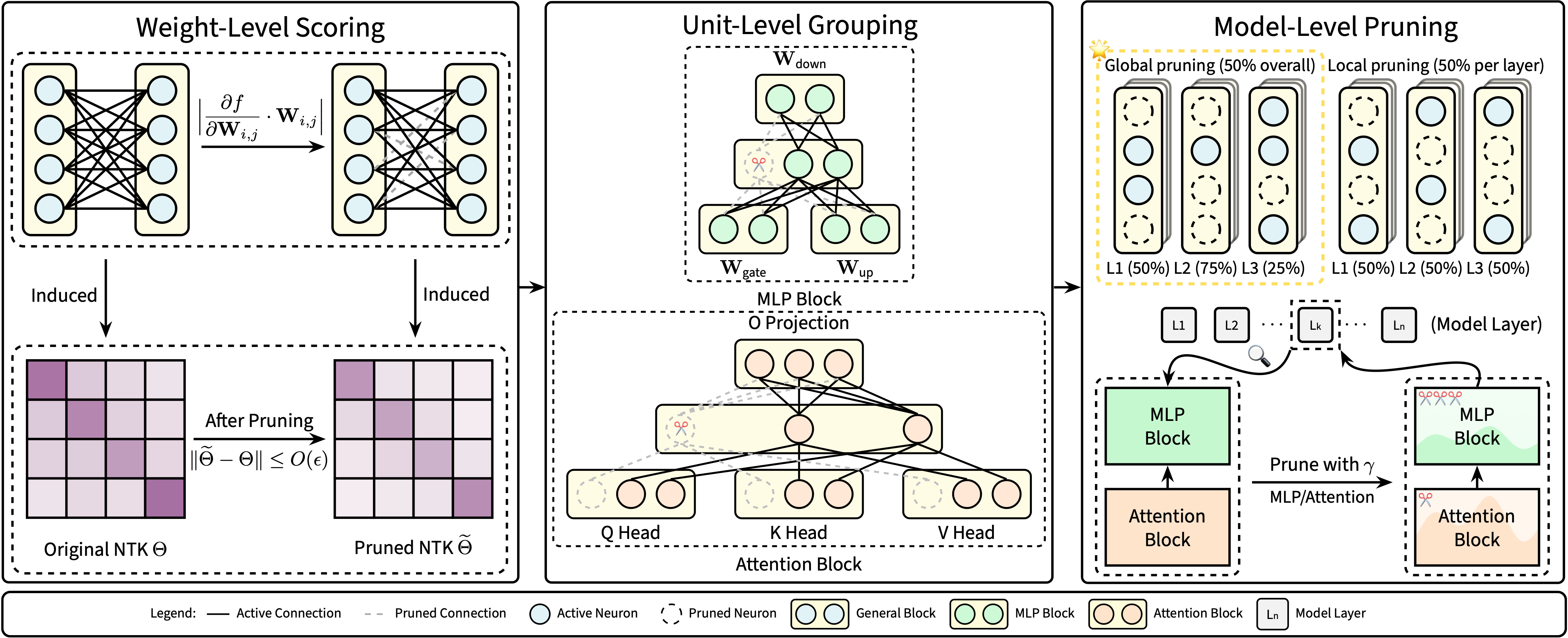}
  \vspace{-0.1in}
  \caption{Illustration of our proposed \textbf{\alg}~framework.
Left: We compute weight-level saliency scores using the NTK-guided score.
Middle: We aggregate the scores into structured units.
Right: We perform global pruning with an adaptive sparsity allocation strategy that adjusts pruning ratios both across layers and between MLP and attention modules with $\gamma$.}
  \label{fig:tasp}
  \vspace{-0.2in}
\end{figure*}

\section{Proposed Method: \textsc{NIRVANA}}
\label{sec:method}

In this section, we present the details of \alg. The corresponding pseudocode is provided in \cref{alg:pruning_group}, and an illustrative diagram is shown in \cref{fig:tasp}.
Our method is built around four core components, each addressing a different level of the pruning process:
\begin{itemize}[itemsep=0pt, parsep=1pt, topsep=0pt]
    \item \textbf{Weight-level saliency scoring:} quantifying the importance of individual weights based on their impact on the model's output (\cref{sec:weight}).
    \item \textbf{Neuron/head-level structured grouping:} aggregating weight saliencies into group-wise scores for MLP neurons and attention heads to enable structured pruning (\cref{sec:unit}).
    \item \textbf{Model-level adaptive sparsity allocation:} applying a global sparsity strategy and balancing pruning between MLP and attention via the adaptive ratio \(\gamma\) (\cref{sec:model}).
    \item \textbf{Calibration data selection:} identifying high-quality calibration data that provides the most informative gradients for reliable pruning (\cref{sec:data}).
\end{itemize}

\subsection{Saliency Score and Its Connection to NTK - Weight Level}
\label{sec:weight}

We begin by deriving the saliency score,
quantifying the \emph{importance} of a weight in terms of its direct impact on the model output, i.e. $\mathcal{I}= f(\mathbf{x}; \mathbf{W})$, 
where \(f(\mathbf{x}; \mathbf{W})\) denote the network output for input \(\mathbf{x}\) and weights \(\mathbf{W}\). Consider removing a single weight \(\mathbf{W}_{i,j}\) by setting it to zero while keeping all other weights unchanged. We define the perturbation as: $\Delta \mathbf{W}_{i,j} = -\mathbf{W}_{i,j},$ 
so that the new weight matrix becomes: $\mathbf{W}' = \mathbf{W} + \Delta \mathbf{W},$ where \(\Delta \mathbf{W}\) is zero everywhere except at entry \((i,j)\). The resulting change in output is: $\Delta f = f(\mathbf{x}; \mathbf{W}') - f(\mathbf{x}; \mathbf{W}).$ Applying a first-order Taylor expansion around \(\mathbf{W}\), we obtain: $f(\mathbf{x}; \mathbf{W} + \Delta \mathbf{W}) \approx f(\mathbf{x}; \mathbf{W}) + \frac{\partial f}{\partial \mathbf{W}_{i,j}} \Delta \mathbf{W}_{i,j},$ which is simplified to:
\vspace{-0.15in}
\begin{small}
\begin{equation*}
    \Delta f \approx -\frac{\partial f(\mathbf{x}; \mathbf{W})}{\partial \mathbf{W}_{i,j}} \mathbf{W}_{i,j}.
\end{equation*}
\end{small}The change in output when pruning \(\mathbf{W}_{i,j}\) is approximately proportional to its gradient scaled by its magnitude. We define the saliency score as the absolute value of this first-order effect:
\begin{small}
\begin{equation}
\label{eq:saliency_score}
{S}(\mathbf{W}_{i,j}) =
\left|
  \frac{\partial f(\mathbf{x}; \mathbf{W})}{\partial \mathbf{W}_{i,j}}
  \cdot
  \mathbf{W}_{i,j}
\right|.
\end{equation}
\end{small}This score quantifies the expected change in model output induced by pruning a weight, and thus serves as our pruning criterion. For next-token prediction, we instantiate the scalar surrogate $f(\mtx x; \mtx W)$ as the mean of the maximum logits over the output sequence. We use this output-based formulation because our analysis targets perturbations to the model function rather than the loss. This distinguishes NIRVANA from loss-gradient pruning methods such as SNIP and makes the criterion less tied to a specific supervised objective.

\noindent\textbf{Bridging to the training dynamics via NTK.}
While the saliency score in \cref{eq:saliency_score} captures the immediate first-order change in model outputs caused by pruning, it does not by itself explain how pruning affects subsequent fine-tuning. Ideally, pruning should not only limit immediate degradation but also preserve the model's trainability after compression. To provide insight into this aspect, we turn to the NTK, which characterizes how model predictions evolve under gradient-based updates. From this perspective, if pruning preserves the NTK, the pruned model is expected to follow a similar optimization trajectory during fine-tuning, suggesting its performance can be recovered efficiently.
Recall the NTK for input \(\mathbf{x}\) is defined as:
\begin{small}
\begin{equation*}
\Theta(\mathbf{x}, \mathbf{x}) = \left\langle
  \nabla_{\mathbf{W}} f(\mathbf{x}; \mathbf{W}),
  \mathrm{sign}\bigl(\nabla_{\mathbf{W}} f(\mathbf{x}; \mathbf{W})\bigr)
\right\rangle,
\end{equation*}
\end{small}where the gradient \(\nabla_{\mathbf{W}} f(\mathbf{x}; \mathbf{W})\) reflects the sensitivity of the model output with respect to each parameter.
For a given weight \(\mathbf{W}_{i,j}\), its contribution to the NTK is proportional to \(\bigl|\partial f / \partial \mathbf{W}_{i,j}\bigr|\).
However, relying solely on sensitivity is insufficient to assess the true impact of pruning, as it ignores the current contribution of the weight itself to the output.
Consider two weights: \(\mathbf{W}_{i,j}\) with a large magnitude but a small gradient, and \(\mathbf{W}_{i',j'}\) with a small magnitude but a large gradient.
Pruning \(\mathbf{W}_{i,j}\) may cause significant output distortion due to its large weight, despite its small gradient; in contrast, pruning \(\mathbf{W}_{i',j'}\) would induce substantial changes in the NTK, as the gradient term dominates, even if its immediate contribution to the output is minor.
This illustrates the need to balance both the parameter value and its sensitivity to avoid disrupting either the output or the training dynamics captured by the NTK.
Accordingly, our saliency score incorporates both aspects, ensuring that pruning decisions simultaneously minimize output perturbations and preserve NTK stability.
We formalize this with the following NTK stability bound:

\begin{proposition}[Short version of Proposition~\ref{prop:ntk_proof}]
\label{prop:ntk}
Let $\mathcal{P}$ denote the set of pruned structured units, where each
unit $g \in \mathcal{P}$ corresponds to a disjoint parameter block
$W_g$. We define the group saliency and its effective gradient magnitude
as
\begin{equation}
S_g
=
\sum_{j \in g}
\left|
\frac{\partial f(X;W)}{\partial W_j} W_j
\right|,
\qquad
\bar{\delta}_g
=
\frac{S_g}{\|W_g\|_1}.
\end{equation}
Assume that the Hessian is locally bounded by $L_H$ along the line
segment between the dense and pruned parameters, and that the pruned
active groups satisfy
\begin{equation}
\delta_{\mathcal{P}}
:=
\min_{g \in \mathcal{P}} \bar{\delta}_g
> 0.
\end{equation}
If the total saliency of the pruned units satisfies
$\sum_{g \in \mathcal{P}} S_g \leq \epsilon$, then
\begin{equation}
\left|
\widetilde{\Theta}(X,X)-\Theta(X,X)
\right|
\leq
\frac{\sqrt{D}L_H}{\delta_{\mathcal{P}}}\epsilon
=
O(\epsilon),
\end{equation}
where $D$ is the parameter dimension of the evaluated structured
layers.
\end{proposition}

This result justifies that pruning based on the saliency score in \eqref{eq:saliency_score} preserves both immediate model outputs (via first-order Taylor approximation) and long-term fine-tuning potential (via NTK stability).
As a result, \alg~achieves both strong zero-shot approximation and robust fine-tuning recovery, as illustrated in \cref{fig:tasp} (left).

\subsection{Structured Pruning via Grouping - Neuron/Attention Head Level}
\label{sec:unit}
To achieve practical efficiency through structured pruning, we group weight saliency scores by their corresponding hidden units. This allows us to prune entire units at once, directly reducing the model's computational dimensions in a strategy that aligns with dependency graph approaches in \citep{Fang_2023_CVPR}.
In the MLP sub-layer, we treat each hidden unit \(u \in \{1,\dots,m\}\) as a group, which consists of all weights in the \(u\)-th column of \(\mathbf{W}_\text{gate}\) and \(\mathbf{W}_\text{up}\), and the \(u\)-th row of \(\mathbf{W}_\text{down}\). For each hidden unit, we compute a cumulative saliency score by summing the saliency scores of its associated weights:
\begin{equation}
\label{eq:mlp_saliency}
\resizebox{\linewidth}{!}{$
\begin{aligned}
S(u)
&=
\sum_{i=1}^{d}\left|\dfrac{\partial f}{\partial (\mathbf{W}_\text{gate})_{i,u}} \cdot (\mathbf{W}_\text{gate})_{i,u}\right|
+\sum_{i=1}^{d}\left|\dfrac{\partial f}{\partial (\mathbf{W}_\text{up})_{i,u}} \cdot (\mathbf{W}_\text{up})_{i,u}\right|
+\sum_{i=1}^{d}\left|\dfrac{\partial f}{\partial (\mathbf{W}_\text{down})_{u,i}} \cdot (\mathbf{W}_\text{down})_{u,i}\right|.
\notag
\end{aligned}
$}
\end{equation}

For MHA, we aggregate the scores for each attention head $a$, which consists of all weights in $\mathbf{W}_a^{Q}, \mathbf{W}_a^{K}, \mathbf{W}_a^{V}$ and the corresponding part $\mathbf{W}^O[a]\in \mathbb{R}^{d_h \times d}$:
\begin{equation}
\label{eq:attn_saliency}
\resizebox{\linewidth}{!}{$
\begin{aligned}
S(a) &=
\sum_{i=1}^{d}\sum_{j=1}^{d_h}
\left|
\dfrac{\partial f}{\partial (\mathbf{W}_a^{Q})_{i,j}}
\cdot
(\mathbf{W}_a^{Q})_{i,j}
\right| 
+
\sum_{i=1}^{d}\sum_{j=1}^{d_h}
\left|
\dfrac{\partial f}{\partial (\mathbf{W}_a^{K})_{i,j}}
\cdot
(\mathbf{W}_a^{K})_{i,j}
\right| 
&+
\sum_{i=1}^{d}\sum_{j=1}^{d_h}
\left|
\dfrac{\partial f}{\partial (\mathbf{W}_a^{V})_{i,j}}
\cdot
(\mathbf{W}_a^{V})_{i,j}
\right| 
+
\sum_{i=1}^{d_h} \sum_{j=1}^d \left|\frac{\partial f}{\partial (\mathbf{W}^O[a])_{i,j}} \cdot (\mathbf{W}^O[a])_{i,j}\right|
\notag
\end{aligned}
$}
\end{equation}
After computing these scores, we rank all hidden units by their aggregated saliency scores. Units falling below a global threshold (determined by our target sparsity level $v$, detailed further in \cref{sec:model}) are pruned.
Specifically, this involves zeroing out the corresponding column in \(\mathbf{W}_\text{gate}\) and $\mtx {W}_\text{up}$, the corresponding row in \(\mathbf{W}_\text{down}\), and the entire attention head $a$.
This grouping-based pruning ensures that the network's dependency structure is respected and that entire units are removed together, leading to real efficiency gains. 
Additionally, we add a hardware-aware rounding to ensure all remaining dimensions are strictly multiples of 8 (further detailed in Section \ref{sec:efficiency}).
See \cref{fig:tasp} (middle) for a visual illustration.

\subsection{Adaptive Sparsity Allocation between Attention and MLP - Model Level}
\label{sec:model}

Having established the unit-level saliency scores, we now address model-level pruning, focusing on how units are ranked globally and how sparsity is allocated between MLP neurons and attention heads.
Most prior works~\citep{sun2023wanda,pmlr-v202-frantar23a,NEURIPS2023_44956951,slicegpt,slimgpt} adopt local sparsity strategies by applying fixed pruning ratios to each layer or module, implicitly treating all units equally regardless of their functional role.
However, recent studies~\citep{adaptpruner,flap} have also noted the challenge of balancing pruning across modules due to scale mismatches, though their approaches primarily rely on heuristic metric normalization.
In contrast, we adopt a principled global sparsity strategy where all units across layers and modules are ranked jointly by their saliency scores, ensuring the overall sparsity target is achieved.
To prevent layer collapse~\citep{NEURIPS2020_46a4378f,JMLR:v24:22-0415}, we introduce a safeguard that retains at least one unit per layer.
Furthermore, motivated by findings that MLP layers store factual knowledge more efficiently than attention heads~\citep{nichani2024understanding}, we introduce a sparsity allocation parameter $\gamma$ to balance their relative pruning rates. We analytically derive the optimal $\gamma$ by quantifying the expected structural influence of each module (detailed in \cref{app:gamma}), ensuring a theoretically principled allocation:
\begin{equation}
    v_{\mathrm{Attn}} = \frac{v (\#_{\mathrm{MLP}} + \#_{\mathrm{Attn}})}{\#_{\mathrm{Attn}} + \gamma \cdot \#_{\mathrm{MLP}}}, \quad
v_{\mathrm{MLP}} = \gamma \cdot v_{\mathrm{Attn}},
\label{eq:gamma}
\end{equation}
where \(v\) denotes the overall target sparsity, \(v_{\mathrm{MLP}}\) and \(v_{\mathrm{Attn}}\) denote the applied sparsity to MLP neurons and attention heads, respectively, and \(\#_{\mathrm{MLP}}\) and \(\#_{\mathrm{Attn}}\) represent their total parameter counts. 
This global ranking and adaptive allocation process is illustrated in \cref{fig:tasp} (right).

\subsection{Calibration Data Selection via KL Divergence}
\label{sec:data}

As pruning transitions from individual weights as in \cref{eq:saliency_score} to larger groups (neurons/heads), the variance in the aggregated saliency scores across these groups can be more pronounced compared to weight-level pruning. This makes it critical to carefully select calibration data that accurately reflects the model's behavior.
Contrary to existing work \citep{Williams_2024,bandari2024c4datasetoptimalpruning,ji2024bewarecalibrationdatapruning}, we empirically find that metrics like data quality, diversity, or quantity do not consistently correlate with post-pruning performance, as illustrated by examples in \cref{app:data_selection}. Therefore, we adopt a lightweight yet effective approach.
We propose using the KL divergence to measure the output discrepancy between the pruned model $\hat{f}$ and the original model $f$, serving as a proxy for selecting calibration data. 
Minimizing KL divergence ensures calibration data induces pruning decisions that preserve the original model's output distribution, avoiding biased gradients from unrepresentative samples. As detailed in \cref{app:kl_justification}, this proxy could be estimated as a Monte Carlo Approximation. 
We define the calibration dataset $\mathcal{C}^{*}$ as:
\[
\mathcal{C}^{*} = \argmin_{\substack{\mathcal{C} \subset \mathcal{D} \\ |\mathcal{C}| = n}}
\frac{1}{n} \sum_{\mathbf{x} \in \mathcal{C}}
\text{KL}\bigl(f(\mathbf{x}) \,\|\, \hat{f}(\mathbf{x})\bigr).
\]
In detail, we randomly sample multiple candidate batches from the dataset and prune the same original dense model using each batch. For each pruned model, we compute the KL divergence on a fixed held-out evaluation set. The batch that yields the lowest KL divergence is the calibration data \(\mathcal{C}^*\). 
This selection process is highly efficient, as each trial requires only a single backward pass on a small data subset, adding minimal computational overhead.
The pseudocode of this process can be found in \cref{alg:data_selection}.

\section{Experiments}
\label{sec:experiments}

\subsection{Experimental Settings}
\label{app:experiment_setup_main}
\textbf{(1) Models and Baselines.}
We conduct our main experiments on \textbf{Llama3.1-8B}, with additional evaluations on Llama3.2-3B, Qwen2.5-7B, Qwen2.5-14B, and T5-base.
We compare \alg~against representative structured pruning methods whose official codebases are publicly available: LLM-Pruner~\citep{NEURIPS2023_44956951}, SliceGPT~\citep{slicegpt}, FLAP~\citep{flap}, and Tyr-the-Pruner \citep{li2025tyrtheprunerstructuralpruningllms}. The results on Olica can be found in \cref{app:exp_olica}.
\textbf{(2) Evaluation Protocol.}
We assess model performance across three dimensions: (1) zero-shot perplexity on standard corpora; (2) zero-shot accuracy on commonsense reasoning and math tasks; and (3) code generation tasks.
All experiments use NVIDIA A100 GPUs.
Please refer to \cref{app:experiment_setup_details} for the complete experimental setup.

\begin{table*}[t]
    \centering
    \resizebox{0.95\linewidth}{!}{
    \begin{tabular}{ll|ccc|ccc|cc|c|l}
        \toprule
        \toprule
        Sparsity& Method & WikiT $\downarrow$ & PTB $\downarrow$ & LambD $\downarrow$ & ARC-e & WinoG &HellaS&SVAMP&MBPP$^\star$ & Avg Acc  & \#Param\\
        \midrule        
        \multirow{1}{*}{0\%} & Llama3.1-8B & 8.50 & 14.02 & 20.09 & 81.52 & 73.56 &78.90&72.67&48.60&71.05&8.03B \\
        \midrule
        \midrule
        \multirow{5}{*}{\parbox{1cm}{20\%}} & LLM-Pruner & \underline{17.45} & \underline{27.89} & \underline{28.33} & \textbf{69.57} & 66.06  &65.89&\underline{22.33}&\underline{4.40}&\underline{45.65}& 6.73B \\
        & SliceGPT & 21.10 & 118.79 & 252.46 & 53.70 & 61.61 &51.40&0.00&0.00&33.34& 6.77B\\
        & FLAP & 20.88 & 31.35 & 31.72 & 60.35 & {66.22} & 59.03 &16.33&3.40 &41.07& 6.61B \\
        & Tyr-the-Pruner & {19.77} & {38.11} & {31.11} & {68.03} & \underline{66.46} & \textbf{67.27} &{19.67}&4.00&{45.09}& 6.66B \\
        & \alg~(ours) & \textbf{13.38} & \textbf{19.77} & \textbf{26.20} & \underline{68.52} & \textbf{67.40} & \underline{66.75} &\textbf{49.00}&\textbf{23.80}&\textbf{55.09}& 6.62B \\
        \midrule
        \multirow{5}{*}{\parbox{1cm}{w/ LoRA$^\dag$}} & LLM-Pruner & \underline{12.57} & \underline{19.77} & \textbf{25.00} & {74.92} & {68.27} & \underline{74.06} &33.33&\underline{25.20}& {55.16}&6.75B \\
        & SliceGPT & 49.81 & 87.25 & {81.64} & 61.53& 61.56 & 59.91 &0.00&0.60&36.72& 6.58B \\
        & FLAP & 16.14& 25.00& {30.63}& 70.71& 63.54& 67.58&{34.33}&15.00&50.23& 6.63B\\
        & Tyr-the-Pruner & {16.65} & {30.15} & \underline{29.68} & \underline{75.67} & \underline{68.51} & {72.40} &\underline{39.67}&21.00&\underline{55.45}& 6.68B \\
        & \alg~(ours) & \textbf{12.37}& \textbf{18.58}& \textbf{25.00}& \textbf{76.98}& \textbf{69.54}&\textbf{74.36}&\textbf{56.67}&\textbf{33.00}& \textbf{62.11}&6.64B \\
        \midrule
        \midrule
        \multirow{5}{*}{\parbox{1cm}{25\%}} & LLM-Pruner & 22.41& 39.33& {33.64}& \underline{60.06}& 56.35&48.16&\underline{16.33}&0.00&36.18& 6.29B\\
        & SliceGPT & \underline{22.19}& 155.54& 280.54& 44.83& 54.15&39.91&0.00&0.00&27.78& 6.57B\\
        & FLAP & 24.71& {37.38} & 34.37& 56.06& \underline{65.27}& {55.60}&11.00&{2.60}&{38.11}& 6.33B\\
        & Tyr-the-Pruner & {22.76} & \underline{31.11} & \underline{31.60} & \textbf{65.03} & {63.85} & \underline{61.82} &{13.33}&\underline{3.20}&\underline{41.45}& 6.35B \\
        & \alg~(ours) & \textbf{15.64} & \textbf{22.41} & \textbf{28.77} & \underline{64.65}& \textbf{65.56}& \textbf{62.21}& \textbf{27.33}& \textbf{16.00}& \textbf{47.15}& 6.28B \\
        \midrule
        \midrule
        \multirow{5}{*}{\parbox{1cm}{33\%}} & LLM-Pruner & \underline{30.15}& \underline{52.92}& \underline{39.94}& {51.85}& 55.80&42.78&\underline{13.67}&0.00&32.82& 5.83B\\
        & SliceGPT & 35.05& 315.42& 446.55& 37.71& 51.62&32.67&0.00&0.00&24.40& 6.06B\\
        & FLAP &  35.39& 54.60& 43.11& 46.00& \underline{60.54}& {48.48}&10.33&0.00&{33.07}& 5.73B\\
        & Tyr-the-Pruner & 30.63 & 54.60 & 43.19 & \underline{55.51} & 56.75  & \underline{52.74} &7.67&0.00&\underline{34.53}& 5.87B \\
        & \alg~(ours) & \textbf{20.40} & \textbf{29.22} & \textbf{34.70} & \textbf{56.44}& \textbf{60.77}& \textbf{53.93}& \textbf{20.67}& \textbf{7.20}& \textbf{39.80}& 5.70B \\
        \midrule
        \midrule
        \multirow{5}{*}{\parbox{1cm}{50\%}}& LLM-Pruner & 215.94 & {356.02} & {196.62} & 31.19 & 49.01 & 28.82 &7.33&0.00  &23.27& 4.55B \\
        & SliceGPT & {93.24} & 612.76 &  870.90 & {32.37} & {49.57} &{29.70}&0.00&0.00&22.33& 4.57B \\
         & FLAP & \underline{68.21} & \underline{97.71} & \textbf{61.87} & {36.45} & \underline{53.56} & \underline{37.20} &\underline{8.33}&0.00&\underline{27.11}& 4.50B \\
         & Tyr-the-Pruner & {123.04} & {196.62} & {115.58} & \textbf{38.64} & {50.59} & {35.39} &{5.33}&0.00&{25.99}& 4.72B \\
        & \alg~(ours) & \textbf{48.94}& \textbf{70.11}& \underline{67.95}& \underline{37.54}&\textbf{54.38}&  \textbf{37.72}&\textbf{9.33}&\textbf{0.20}&\textbf{27.83}& 4.51B \\        
        \midrule
        \multirow{5}{*}{\parbox{1cm}{w/ LoRA$^\dag$}} & LLM-Pruner & 45.26& 92.87& 70.12& {48.32} & 52.41 & {43.51}&4.67&0.00&29.78& 4.55B \\
        & SliceGPT & 198.93 & 246.61 &  248.55 & 40.19 & {52.49} &36.57 &0.00&0.00&25.85& 4.59B \\
        & FLAP & \underline{28.33}& \underline{42.52}& \underline{43.87}& {50.76}& \textbf{58.09}& \underline{49.90}&\underline{15.67}&0.00&\underline{34.88}& 4.52B\\
        & Tyr-the-Pruner & {54.60} & {81.96} & {81.96} & \underline{52.40} & {54.06} & {46.09} &{13.33}&0.00&{33.18}& 4.74B \\
        & \alg~(ours) & \textbf{25.79}& \textbf{36.94}& \textbf{42.52}& \textbf{57.49} & \underline{56.27} & \textbf{49.91} &\textbf{23.00}&\textbf{3.40}&\textbf{38.01} &4.53B \\
        \bottomrule
        \bottomrule
    \end{tabular}
    }
    \label{tab:llama}
    \vskip -0.15in
    \caption*{\raggedright\tiny\textsuperscript{$\quad\star$} Pass@1. 3-shot.\raggedright\tiny\textsuperscript{$\quad\dag$} We report the results at representative sparsity levels. }
    \vskip -0.05in
    \caption{Zero-shot and recovery fine-tuning evaluation results of Llama3.1-8B. \textbf{Bold} indicates the best results while \underline{underline} indicates the second-best. The down-arrow notation (\textdownarrow) indicates that a lower metric represents better performance.
    }
    \vskip -0.25in
\end{table*}

\subsection{Results on Llama3.1-8B}

\cref{tab:llama} summarizes the evaluation results of Llama3.1-8B across various sparsity levels and benchmarks. Additional experiments demonstrating the consistent effectiveness of our method across varying model sizes (Llama3.2-3B), model architectures (T5-base), and model families (Qwen) can be found in \cref{app:experiment}. Overall, \alg~consistently establishes the state-of-the-art across both zero-shot and recovery fine-tuning settings.

\textbf{Preservation of Complex Reasoning Capabilities.} Structured pruning unavoidably inflicts degradation on zero-shot performance. While most baselines manage to retain the language modeling capability and commonsense reasoning, they suffer from a collapse on more challenging, logic-heavy tasks. For instance, on the code generation task (MBPP) and math word problems (SVAMP), baselines plummet to near-zero performance even at low sparsity (20\%). In stark contrast, \alg~exhibits remarkable robustness, outperforming the best baseline on MBPP by a massive margin ($23.80$ vs $4.40$). This substantial gap highlights a critical insight: \alg's explicit module-aware allocation and global NTK-guided output-space scoring successfully identify and protect the critical structural units responsible for the model's complex reasoning and generative capabilities. 

\textbf{Superior Fine-tuning Recoverability.} A practical pruning framework should not only maintain zero-shot fidelity but also preserve the model's optimization landscape for efficient recovery. Under the lightweight LoRA fine-tuning setting, \alg~achieves the highest recovered accuracy across all sparsity levels. This validates our theoretical motivation: aligning the pruning criterion with the NTK proxy ensures that the essential training dynamics are preserved, making the pruned model highly amenable to downstream tuning.

We observe limitations in existing approaches: (1) \textit{LLM-Pruner} performs reasonably well at 20\% sparsity but degrades rapidly at higher compression rates. (2) \textit{SliceGPT} exhibits extreme sensitivity to calibration data, leading to brittle and unstable generative performance. (3) \textit{FLAP} yields relatively better results at higher sparsity but sub-optimal at moderate levels. (4) \textit{Tyr-the-Pruner} effectively preserves general reasoning task abilities but suffers a collapse on harder generation tasks. 
By contrast, \alg~delivers better performance across all metrics and compression regimes. We also provide qualitative analysis in \cref{app:sample}.

\begin{figure}[t!]
  \centering
  \vspace{-0.1in}
  \includegraphics[width=0.9\linewidth, trim=0.4cm 0.7cm 0.5cm 0.5cm, clip]{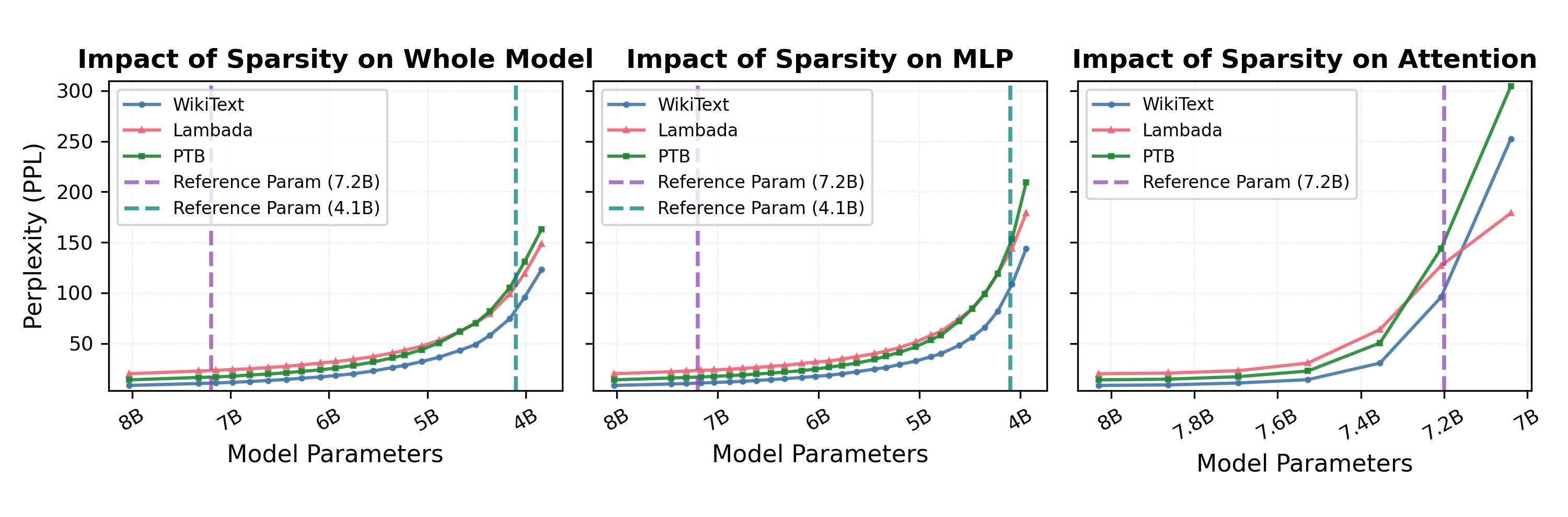}
  \vspace{-0.15in}
  \caption{Comparison of pruning impact across different pruning scopes.
We evaluate pruning applied to the entire model, MLP modules only, and attention heads only. 
Dashed vertical lines indicate reference parameter points (7.2B and 4.1B) for visual comparison.
}
  \label{fig:sparsity_ratio}
\vspace{-0.25in}
\end{figure}

\begin{table}[t]
\centering
\vspace{0.2in}
\resizebox{\textwidth}{!}{
\begin{tabular}{lccccccccc}
\toprule
Method & WikiT $\downarrow$ & PTB $\downarrow$ & LambD $\downarrow$ & ARC-e & WinoG & HellaS & SVAMP & MBPP & Avg Acc \\
\midrule
Loss & 18.87 & 23.12 & 31.11 & \textbf{70.79} & 65.75 & 65.97 & 43.33 & 10.80 & 51.33 \\
NIRVANA (output) & \textbf{13.38} & \textbf{19.77} & \textbf{26.20} & 68.52 & \textbf{67.40} & \textbf{66.75} & \textbf{49.00} & \textbf{23.80} & \textbf{55.09} \\
\bottomrule
\end{tabular}
}
\vspace{-0.1in}
\caption{Ablation study on Llama3.1-8B on loss vs output gradients.}
\label{tab:ablation_loss}
\vspace{-0.2in}
\end{table}

\subsection{Ablation Study \& Hyper Parameter Analysis}

\begin{wraptable}{r}{0.45\linewidth}
\centering
\vspace{-0.15in}
\resizebox{\linewidth}{!}{
\begin{tabular}{lccc}
\toprule
\toprule
     & Wikitext  & PTB & Lambada\\
\midrule
\alg &  {48.94} & {70.11} & {67.95}\\
\midrule
Magnitude score  & $\approx\text{10}^\text{6}$ & $\approx\text{10}^\text{5}$ & $\approx\text{10}^\text{6}$\\
NTK-SAP score  &58.12	&72.33	&72.33 \\
local & 90.01 & 142.85 & 132.42\\
w/o ratio $\gamma$& 102.00 & 139.42 & 123.04 \\
w/o selected data & 72.33 & 115.58 & 95.82 \\
\bottomrule
\bottomrule
\end{tabular}
}
\vskip -0.05in
\caption{Ablation study on Llama3.1-8B with 50\% sparsity on three PPL datasets.}
\vspace{-0.15in}
\label{tab:ab_study}
\end{wraptable}
Our method includes several key components: (1) an NTK-inspired saliency score; (2) an adaptive sparsity allocation strategy that combines global sparsity ranking with the pruning ratio \(\gamma\) to balance MLP units and attention heads; and (3) a calibration data selection strategy.
We conduct an ablation study to assess the effects of these components, as shown in \cref{tab:ab_study}.
Specifically, we compare \alg\ with magnitude scoring, saliency score from NTK-SAP 
\citep{wang2023ntksapimprovingneuralnetwork}, local pruning, without different ratio between MLP and attention 
($\gamma=1$), and without KL-selected calibration data.
We find that magnitude-based scores lead to extreme performance collapse, showing the inadequacy of naive importance metrics in LLM pruning.
Similarly, applying local pruning instead of global pruning also leads to a performance drop, likely due to its inability to account for cross-layer importance differences. Additionally, since \alg\ is different than the traditional methods on the gradient calculation, we provide further ablation on this choice in \cref{tab:ablation_loss}.

\textbf{Impact of $\gamma$.} To better understand the impact of $\gamma$, we further investigate the impact of pruning scope by analyzing how perplexity changes as a function of the remaining model parameters.
\cref{fig:sparsity_ratio} presents a detailed comparison where pruning is applied only to attention heads, only to MLP units, or jointly across the whole model with $\gamma$.
We observed that performance degrades non-linearly as the model is pruned. Targeting only attention heads leads to a catastrophic performance collapse even at moderate sparsity. While only pruning MLP is more stable, performance still drops sharply below a critical threshold (4.1B).
In contrast, NIRVANA yields a smoother perplexity curve and consistently outperforms both specialized approaches at any given parameter budget. This demonstrates that balancing pruning across different components is crucial for maintaining model robustness.

\begin{figure}[t]
\vspace{-0.2in}
    \centering
    \includegraphics[width=0.5\linewidth,trim=0.35cm 0 0.22cm 0, clip]{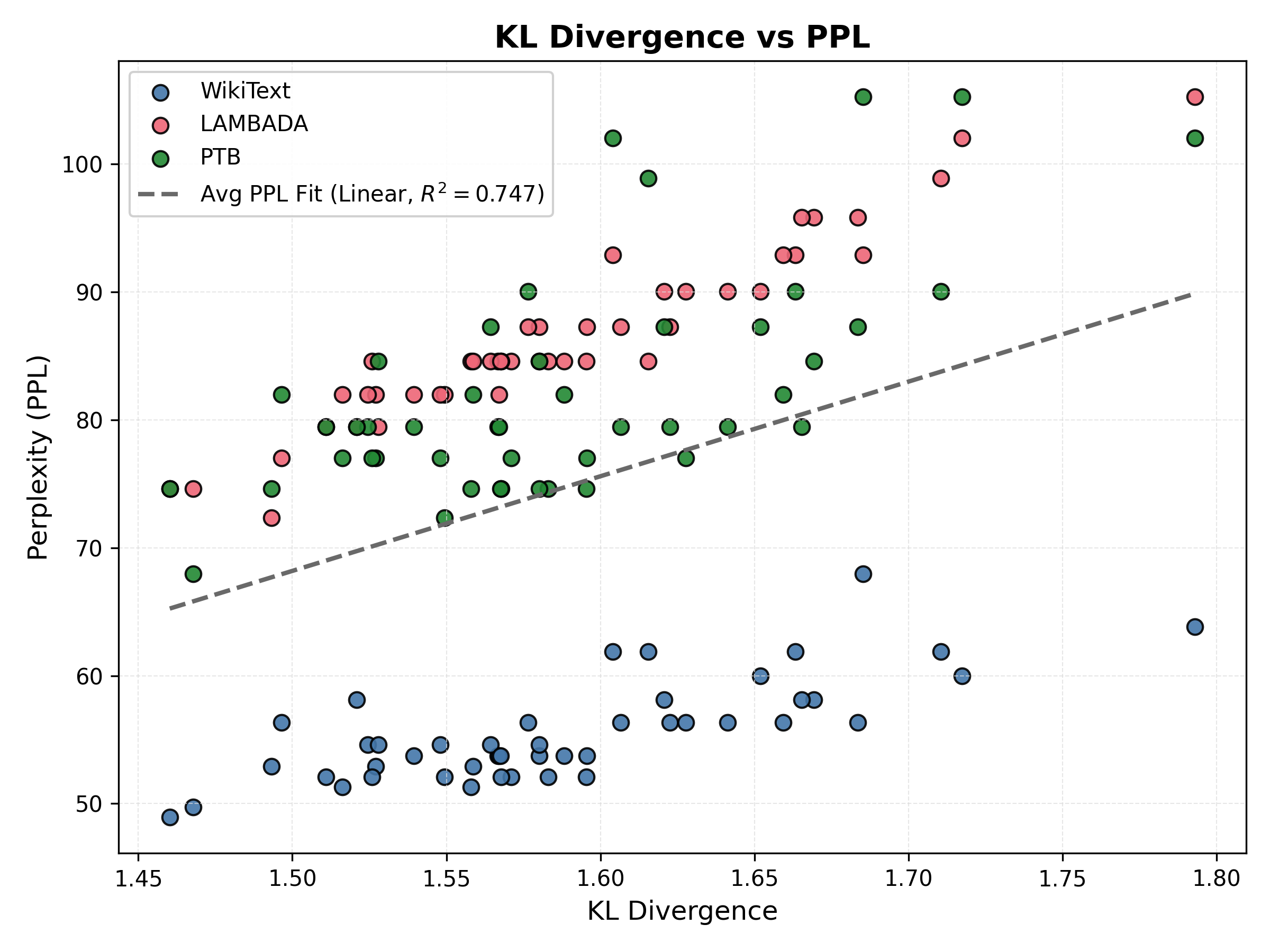} 
    \vspace{-0.1in}
    \caption{Relationship between KL divergence and perplexity (PPL) across different datasets over 50 runs. The observed linear correlation supports the effectiveness of using KL divergence as a proxy for data selection in pruning calibration.}
    \label{fig:kl_data_select}
\vspace{-0.1in}
\end{figure}

\textbf{Calibration Data Selection.}
We use calibration data (32 examples, 128 tokens each) from BookCorpus. To identify the optimal subset, we sample multiple candidate batches and compute the KL divergence between the original and pruned models on a fixed validation set. Since this KL proxy strictly correlates with downstream zero-shot perplexity (see \cref{fig:kl_data_select}), we select the calibration batch that minimizes the KL divergence for all subsequent experiments. Detailed analyses are deferred to \cref{app:kl_justification,app:data_selection}.

\subsection{Efficiency Analysis}
\label{sec:efficiency}

\begin{wraptable}{r}{0.6\linewidth}
\vspace{-0.15in}
\centering
\resizebox{\linewidth}{!}{
\begin{tabular}{llccccc}
\toprule
\toprule
  Spsty & Method &\#Param &FLOPs & Lat (s/bc) $\downarrow$&	TPT (ktps) $\uparrow$ & Mem (GB) \\
\midrule
0\% & Llama3.1-8B& 8.03B & 6.98G & 0.33 & 12.41 & 17.95 \\
\midrule
\multirow{4}{*}{20\%} & LLM-Pruner & 6.73B & 5.75G & 0.41 & 9.99 & 15.46 \\
                    & SliceGPT & 6.57B&	5.59G	&0.38&	10.74	&15.54 \\
                    & FLAP & 6.61B & 5.56G & 0.50 & 8.19 & 15.33 \\
                      & NIRVANA & 6.62B & 5.61G & 0.28 & 14.63 & 15.29 \\
\midrule
\multirow{4}{*}{50\%} & LLM-Pruner & 4.55B & 3.76G & 0.20 & 20.48 & 10.95 \\
                      & SliceGPT & 4.57B	&3.52G&	0.27&	15.34 &	10.53 \\
                      & FLAP & 4.50B & 3.52G & 0.30 & 13.65 & 11.32 \\
                      & NIRVANA & 4.51B & 3.55G & 0.21 & 19.50 & 11.27 \\
\bottomrule
\bottomrule
\end{tabular}
}
\vspace{-0.1in}
\caption{Post-pruning statistics during inference, tested on 20 rounds of Wikitext (batch size 32).}
\vspace{-0.15in}
\label{tab:efficiency}
\end{wraptable}
\textbf{Inference Efficiency.} 
To evaluate the practical performance gains of structured pruning, we benchmark the inference efficiency of all methods on a single A100. The results are presented in Table~\ref{tab:efficiency}. 
A key observation is the stark difference in how parameter reduction translates into practical speedups. While all methods successfully reduce peak memory usage, their impact on latency and throughput varies significantly. \alg~demonstrates a consistent trend: as sparsity increases, latency steadily decreases, and throughput improves accordingly.
Baseline methods, however, exhibit erratic and non-monotonic behavior. For instance, LLM-Pruner's latency unexpectedly deteriorates at 20\% sparsity, despite the reduction in parameters. This suggests that merely reducing FLOPs does not guarantee acceleration.
These inconsistent speedups are caused by a hardware-software mismatch. Modern GPUs leverage specialized compute kernels (e.g., Tensor Cores) that are highly optimized for matrix dimensions that are \textbf{multiples of 8} \citep{nvidia_tensor_cores_2024,nvidia_mixed_precision_docs}. When pruning creates dimensions not aligned to this constraint, the GPU falls back to slower, general-purpose kernels, which can negate the performance gains from sparsity. See \cref{app:efficiency} for the detailed analysis.
Armed with this insight, we designed \alg~to be explicitly \textit{hardware-aware}, which includes a dimension alignment step that ensures all remaining hidden dimensions are multiples of 8. For the trade-off between quantization and between unstructured/semi-structured pruning, please refer to \cref{app:quantization} and \cref{app:up_sp}.

\textbf{Computational cost of the KL-based Selection.}
We measured the actual wall-clock time for the KL-based selection process on a single A100 GPU. As shown in \cref{tab:kl_cost}, the cost is minimal.
This low latency is achieved because we use a small calibration set. As discussed in \cref{app:data_selection}, neither the number of examples nor sequence length dominates pruning quality. Therefore, we fix the candidate calibration set to 32 samples and the evaluation "anchor set" to 128 samples. The process consists of two steps: 
\textbf{1. Pre-computation (Once):} 
\begin{wraptable}{l}{0.4\linewidth}
\centering
\resizebox{0.9\linewidth}{!}{
\begin{tabular}{lc}
\toprule
\toprule
  Model   &  Wallclock time  \\
\midrule
Llama3.1-8B&	4.12s \\
Qwen2.5-14B&	6.28s \\
\bottomrule
\bottomrule
\end{tabular}    
}
\caption{Data selection time.}
\label{tab:kl_cost}
\end{wraptable}Computing dense model logits on the anchor set takes 3.04s for Llama3.1-8B and 3.93s for Qwen2.5-14B on a single A100. This is cached and reused. 
\textbf{2. Selection Loop (Per Candidate):} For each candidate, computing the pruned model's logits and the KL divergence takes 4.12s (Llama3.1-8B) and 6.28s (Qwen2.5-14B).
Combined with the fast saliency computation (approx. 2s, detailed in \cref{app:efficiency}), a complete search over 50 candidate sets for Llama3.1-8B takes less than 10 minutes on a single A100. This minimal one-time cost yields significant performance gains, making it a highly efficient trade-off.

\section{Conclusion}
\label{sec:conclusion}

We introduced \textbf{NIRVANA}, a novel structured pruning approach guided by the NTK, aiming to balance zero-shot accuracy preservation with fine-tuning adaptability.
By explicitly linking the pruning criterion to NTK dynamics, NIRVANA ensures pruning decisions that maintain both immediate output stability and fine-tuning recovery.
Furthermore, our method integrates an adaptive sparsity allocation mechanism and a KL-divergence-based calibration data selection strategy, enhancing pruning robustness and efficiency.
Extensive experiments demonstrate that NIRVANA consistently outperforms existing structured pruning baselines in downstream task performance under equivalent sparsity. Although effective, \alg\ does have several limitations: similar to all one-shot pruning methods, the performance drop at high sparsity levels could be more severe, and further substantial re-training would be required for the recovery of the performance under such circumstances.

\newpage
\clearpage

\section*{Acknowledgments}
This work is supported by National Science Foundation under Award No. IIS-2117902. The views and conclusions are those of the authors and should not be interpreted as representing the official policies of the funding agencies or the government.
This work used Delta at NCSA through allocation CIS250202 from the Advanced Cyberinfrastructure Coordination Ecosystem: Services \& Support (ACCESS) program, which is supported by U.S. National Science Foundation grants \#2138259, \#2138286, \#2138307, \#2137603, and \#2138296.

\section*{Ethical Statement} 
As a model compression method, NIRVANA primarily targets reducing the resource demands of large language models without altering their behavior or capabilities.
We do not foresee direct societal risks or misuse arising uniquely from this work.
However, as with any efficiency technique, care should be taken to ensure that improved accessibility to large models does not inadvertently facilitate harmful or unintended applications.

\section*{Reproducibility statement} 
To ensure our work is fully reproducible, we've provided extensive details on our experimental setup, hardware specifications, and algorithms, complete with pseudocode and illustrative figures in both the main body and in the appendix.
For direct validation and easy adoption, a complete, runnable, and easy-to-integrate code implementation is available in the supplementary materials.

\bibliography{references}
\bibliographystyle{colm2026_conference}

\newpage
\appendix
\onecolumn

\tableofcontents

\addtocontents{toc}{\protect\setcounter{tocdepth}{2}}

\clearpage
\newpage

\section{Related Work}
\label{app:background_pruning}

While 
\cref{sec:problem_formulation} provides a high-level overview of LLM structured pruning; this appendix offers a more comprehensive discussion. The landscape of modern AI is fundamentally shaped by the Transformer architecture \citep{zou2025gtr,chencodesteer,zhang2024knowledge,zhang-etal-2023-vibe,cui2025dr,10.1145/3690624.3709196}, whose scaling capabilities have driven unprecedented progress across different tasks \citep{zou2025transformercopilotlearningmistake,zou2025reasonfluxprmtrajectoryawareprmslong,chen2025r1,He_2025,ning2,ninggraph4mm}. This success, however, has also necessitated a strong focus on computational efficiency to manage the growing resource demands of large models \citep{10.1145/3583780.3615136,zhao2025secondorderfinetuningpainllmsa,zhao2025saberswitchablebalancedtraining,10.1145/3690624.3709213}. Against this backdrop, we now dive into a more detailed analysis of existing pruning methodologies.

\subsection{Notations}
We use lowercase letters to denote scalars, boldface lowercase letters for vectors, and boldface uppercase letters for matrices. The element-wise product is denoted by \(\odot\). Let \(f(\mathbf{x};\mathbf{W}\odot\mathbf{M})\) denote the neural network function, where \(\mathbf{x}\) are the inputs, \(\mathbf{W}\) the weights (or connections), and \(\mathbf{M}\) the sparse mask matrix with sparsity \(v\) (density \(d=1-v\)). Additionally, let \(\mathcal{L}\) be the loss function, and let 
\[
D=\{(\mathbf{x}_{k},\mathbf{y}_{k})\}_{k=1}^{N}\subset \mathbb{R}^{n}\times \mathbb{R}^{m}
\]
denote the dataset with \(N\) data points (with \(\mathbf{x}\) as the input features and \(\mathbf{y}\) as the labels). Finally, let \(\mathcal{A}\) be the optimizer that, given the initial weights of the network before supervised fine-tuning (SFT) \(\mathbf{W}^{(0)}\), returns the weights after SFT, i.e., \(\mathbf{W}^{\text{(final)}}=\mathcal{A}(\mathbf{W}^{(0)})\). See the full notation in \cref{tab:notation}.

\subsection{Post-Train Pruning}

Post-train pruning compresses a fully trained dense model by removing unimportant weights or structures. A common formulation minimizes the discrepancy between the outputs of the uncompressed and pruned layers. Given an input \(\mathbf{x}\), the objective is to solve:
\begin{align*}
&\text{argmin}_{\hat{\mathbf{W}},\mathbf{M}}\mathcal{L}\Bigl(\mathbf{W}^{\text{(final)}}\mathbf{x},\,(\hat{\mathbf{W}}\odot\mathbf{M})\mathbf{x}\Bigr)\\[1mm]
=&\text{argmin}_{\hat{\mathbf{W}},\mathbf{M}}\frac{1}{N}\sum_{k=1}^{N}\mathcal{L}\Bigl(\mathbf{W}^{\text{(final)}}\mathbf{x}_{k},\,(\hat{\mathbf{W}}\odot\mathbf{M})\mathbf{x}_{k}\Bigr),
\end{align*}
where \(\mathbf{M}\) is a mask matrix enforcing a fixed sparsity \(v\).

Directly solving this joint optimization over \(\hat{\mathbf{W}}\) and \(\mathbf{M}\) is NP-hard. Consequently, popular practices include fixing the weights (i.e., setting \(\hat{\mathbf{W}}=\mathbf{W}\)) and searching for \(\mathbf{M}\) only (one-shot pruning \citep{frankle2018the,pmlr-v202-frantar23a,sun2023wanda,NEURIPS2021_a376033f}), or selecting \(\mathbf{M}\) first and then optimizing \(\hat{\mathbf{W}}\) (which typically requires further fine-tuning or re-training \citep{NEURIPS2022_987bed99,NEURIPS2023_44956951}).

\subsection{Foresight Pruning}
\label{app:foresight_pruning}

Foresight pruning, also known as pruning before training, seeks to identify and eliminate redundant parameters at initialization, thereby reducing both training and inference costs. For a neural network \(f\) parameterized by \(\mathbf{W}\) and data \((\mathbf{x},\mathbf{y})\), the objective is formulated as:
\begin{equation}
\label{eq:foresight}
\begin{aligned}
&\min_{\mathbf{M}}\;\mathcal{L}\Bigl(f\bigl(\mathbf{x};\mathcal{A}(\mathbf{W}^{(0)}\odot\mathbf{M})\bigr),\mathbf{y}\Bigr)\\
    =&\min_{\mathbf{M}}\;\frac{1}{N}\sum_{k=1}^{N}\mathcal{L}\Bigl(f\bigl(\mathbf{x}_k;\mathcal{A}(\mathbf{W}^{(0)}\odot\mathbf{M})\bigr),\mathbf{y}_k\Bigr),
\end{aligned}
\end{equation}
where \(\mathcal{A}\) returns the final weights \(\mathbf{W}^{\text{(final)}}\). As in the post-train case, solving \cref{eq:foresight} exactly is NP-hard because it involves joint optimization over the mask and the model parameters.

To bridge this gap, popular approaches define a \emph{saliency score} \(S_{i,j}\) for each weight, which estimates the impact of removing the connection \(\mathbf{W}_{i,j}\). A general form of the saliency score is:
\begin{equation}
S\bigl(\mathbf{W}^{(0)}_{i,j}\bigr) = \frac{\partial \mathcal{I}}{\partial \mathbf{W}^{(0)}_{i,j}} \cdot \mathbf{W}^{(0)}_{i,j},
\end{equation}
where \(\mathcal{I}\) is a function that measures the importance of the weight \(\mathbf{W}\) to the network \(f\). Once these scores are computed, the mask is obtained by selecting the top \(\kappa\%\) of weights:
\[
\mathbf{M}_{i,j} = \text{Top}_{\kappa}(S)_{i,j} = \begin{cases}
1, & \text{if } S_{i,j} \text{ is among the top } \kappa\%,\\[1mm]
0, & \text{otherwise}.
\end{cases}
\]

\begin{table}[t]
\caption{Notation.}
\label{tab:notation}
\vskip 0.15in
\begin{center}
\begin{small}
\begin{tabular}{ll}
\toprule
Symbol & Definition and Description \\
\midrule
$\mathbf{x}$ & Input token, $\mathbf{x} \in \mathbb{R}^{d}$ \\
$f(\mathbf{x}; \mathbf{W} \odot \mathbf{M})$ & Neural network function, with weights $\mathbf{W}$ and sparse mask $\mathbf{M}$ \\
$\mathbf{W}$ & Weight matrices (parameters) of the network \\
$\mathbf{M}$ & Binary mask matrix, $\mathbf{M} \in \{0,1\}^{\text{shape}(\mathbf{W})}$, indicating pruned weights \\
$\mathcal{D}$ & Dataset \\
$S$ & Saliency score function \\
$\odot$ & Element-wise (Hadamard) product \\
$\sigma(\cdot)$ & Swish activation function, applied elementwise \\
$v$ & Sparsity ratio \\
$d$ & Hidden size of the model \\
$m$ & Intermediate (FFN) dimension in the MLP \\
$h$ & Number of attention heads \\
$d_h$ & Per-head dimension, $d_h = d/h$ \\
$\mtx{W}_a^Q, \mtx{W}_a^K, \mtx{W}_a^V$ & Query, Key, Value weight matrices for head $i$, each $\in \mathbb{R}^{d \times d_h}$ \\
$\mtx{W}_a^O$ & Output weight matrices for head $i$, $\in \mathbb{R}^{d_h \times d}$ \\
$\mtx{Q}_a, \mtx{K}_a, \mtx{V}_a$ & Query, Key, Value representations for head $i$ \\
$\mathbf{W}_\text{gate}, \mathbf{W}_\text{up}$ & Projection weights, $\in \mathbb{R}^{d \times m}$ \\
$\mathbf{W}_\text{down}$ & Down projection weight, $\in \mathbb{R}^{m \times d}$ \\
$\text{head}_{a}$ &  Attention output for head $a$ \\
$\text{MHA}(\mtx x)$ & Multi-Head Attention output \\
$\mathbf{H}(\mathbf{x})$ & MLP block output \\
\bottomrule
\end{tabular}
\end{small}
\end{center}
\vskip -0.1in
\end{table}

\subsection{Revisiting Saliency Methods}

Several representative methods use different formulations of the saliency score to approximate weight importance. These approaches are predominantly explored in the context of computer vision tasks. In the following sections, we will also discuss the work in the LLM context.

\textbf{SNIP}~\citep{lee2019snipsingleshotnetworkpruning} proposes a data-dependent saliency:
\[
S_{\text{SNIP}} = \Bigl|\frac{\partial \mathcal{L}(\mathbf{x}; \mathbf{W})}{\partial \mathbf{W}_{i,j}} \cdot \mathbf{W}_{i,j}\Bigr|.
\]

\textbf{GraSP}~\citep{wang2020pickingwinningticketstraining} employs a second-order (Hessian) metric:
\[
S_{\text{GraSP}} = -\Bigl(\mathbf{H}\frac{\partial \mathcal{L}(\mathbf{x}; \mathbf{W})}{\partial \mathbf{W}_{i,j}}\Bigr) \cdot \mathbf{W}_{i,j},
\]
where \(\mathbf{H}\) denotes the Hessian of the loss.

\textbf{SynFlow}~\citep{NEURIPS2020_46a4378f} introduces a data-agnostic approach by defining saliency on constant inputs (e.g., \(\mathbf{1}\)) and absolute weights:
\[
S_{\text{SynFlow}} = \Bigl|\frac{\partial f(\mathbf{1};\,|\mathbf{W}|)}{\partial |\mathbf{W}_{i,j}|}\Bigr| \cdot \bigl|\mathbf{W}_{i,j}\bigr|.
\]
Although SynFlow's formulation is similar to our approach, using absolute values may not fully capture the true gradient flow in the model.

\textbf{NTK-SAP}~\citep{wang2023ntksapimprovingneuralnetwork} adopts a data-agnostic perspective by injecting a small perturbation \(\Delta \mathbf{W}_{i,j}\sim \mathcal{N}(0,\epsilon\,\mathbf{I})\):
\[
S_{\text{NTK-SAP}} = \Bigl|\frac{\partial \,\|f(\mathbf{z}; \mathbf{W}) - f(\mathbf{z}; \mathbf{W}+\Delta\mathbf{W})\|_2^2}{\partial \mathbf{W}_{i,j}}\Bigr|,
\]
with \(\mathbf{z}\) drawn from a standard normal distribution.

\textbf{PX-Pruning}~\citep{Iurada_2024_CVPR} introduces an auxiliary function \(\mathcal{R}\) computed by two helper networks \(g\) and \(h\) (sharing the original architecture):
\[
S_{\text{PX}} = \Bigl|\frac{\partial \,\mathcal{R}(\mathbf{x},\mathbf{W},\mathbf{a})}{\partial (\mathbf{W}_{i,j}^2)} \cdot \mathbf{W}_{i,j}^2\Bigr|,
\]
where
\[
\mathcal{R}(\mathbf{x},\mathbf{W},\mathbf{a}) = g\bigl(\mathbf{x}^2,\mathbf{1},\mathbf{a}\bigr) \, h\bigl(\mathbf{1},\mathbf{W}^2,\mathbf{1}\bigr),
\]
and \(\mathbf{a}\) tracks the activation status. Backpropagation through \(\mathcal{R}\) yields a saliency score for each parameter.

While these foresight methods have shown promise, they are not commonly applied to LLMs due to the models' scale and behavior during fine-tuning. Moreover, the inherent nature of unstructured pruning in these methods often limits their ability to reduce training costs. In contrast, our work focuses on the MLP module in LLMs and develops an NTK-aware saliency score tailored to preserve training dynamics.

\subsection{Detailed Comparison with NTK-SAP}

\begin{itemize}
    \item Paradigm and Target: NTK-SAP is an iterative, unstructured pruning method that finds sparse subnetworks at initialization. In contrast, NIRVANA is a one-shot, structured pruning method designed for post-training compression of LLMs (removing entire heads/neurons).
    \item Saliency Derivation: NTK-SAP aims to preserve the NTK spectrum; its saliency score is a finite-difference approximation of the NTK trace norm: $S(m_{ij}) =\left|\frac{\partial \mathbb{E}[||f(z; W\odot m) - f(z; (W+ \Delta W) \odot m)||_2^2]}{\partial m_{ij}}\right|$, where $m$ represents the mask. NIRVANA's saliency score is derived from a first-order Taylor expansion to directly estimate the impact of weight removal on the model's output: $S(W_{ij})=|\frac{\partial f}{\partial W_{ij}}\cdot W_{ij}|$, where $W$ represents the weight. We then analyze this score under an Adam-style NTK to justify that it preserves NTK stability during optimization.
    \item Data and Weight Dependency: NTK-SAP is data- and weight-agnostic (using random inputs $z$ and averaging over random initializations). NIRVANA is data- and weight-dependent, using real calibration data and specific pre-trained weights to compress a target model instance. This distinction is critical. Data- and weight-agnostic methods are blind to the specific knowledge encoded in pre-trained weights, do not leverage the specific structure of a given pre-trained checkpoint, which can make it overlook model-specific important directions. By conditioning on the actual weights, NIRVANA explicitly targets the preservation of these learned knowledges, whereas agnostic methods risk pruning essential capabilities.
\end{itemize}

\subsection{LLM-based Pruning}

Recent unstructured pruning methods, such as SparseGPT~\citep{pmlr-v202-frantar23a} and Wanda~\citep{sun2023wanda}, have proposed efficient one-shot pruning strategies that remove individual weights based on local reconstruction error or activation-aware criteria.
While effective in preserving accuracy, their irregular sparsity patterns limit practical speedups on existing hardware accelerators.
To bridge the gap between unstructured and structured sparsity, LLM-Barber~\citep{su2025llmbarberblockawarerebuildersparsity} proposes a block-aware rebuilder for semi-structured masks, aiming to balance flexibility and hardware efficiency.

In contrast, structured pruning methods aim to remove entire neurons, attention heads, or layers, enabling more hardware-friendly sparsity.
Representative approaches include LLM-Pruner~\citep{NEURIPS2023_44956951}, Sheared Llama~\citep{xia2024shearedllamaacceleratinglanguage}, and SlimGPT~\citep{slimgpt}, which prune model components based on local importance scores.
More recently, Olica~\citep{he2025olica} and SlimLLM~\citep{guo2025slimllm} have pushed the boundaries of structured pruning by refining importance estimation metrics. However, these methods typically treat attention and MLP modules uniformly or may face challenges with the reduced redundancy in Grouped Query Attention (GQA) architectures.

Several recent works have attempted to address the imbalance across layers.
For example, Adapt-Pruner~\citep{adaptpruner} introduces layer-wise global scoring but still applies uniform pruning within modules.
FLAP~\citep{flap} further introduces a heuristic structure search to assign different sparsity levels across both layers and modules.
Similarly, T{\'y}r-the-Pruner~\citep{li2025tyrtheprunerstructuralpruningllms} optimizes global sparsity distribution via evolutionary search. While effective, such search-based methods can be computationally prohibitive (e.g., taking hours for a single run) compared to one-shot approaches.
ShortGPT~\citep{shortgpt} removes entire layers based on global layer importance, but does not differentiate between module types within layers.

SliceGPT~\citep{slicegpt} instead prunes hidden dimensions rather than intermediate MLP or attention dimensions, leveraging computational invariance in RMSNorm-connected transformers.
The heavy reliance on calibration data to compute these transformations makes it sensitive to the choice of calibration dataset.
\textbf{Pruning with Fine-tuning.} Distinct from the aforementioned post-training pruning methods, another line of work, including LoRAP~\citep{li2024loraptransformersublayersdeserve} and LoRAPrune~\citep{zhang2024loraprunestructuredpruningmeets}, tightly couples structured compression with Low-Rank Adaptation (LoRA) specifically for fine-tuning tasks.
Our work, in contrast, focuses on architecture compression for the general-purpose backbone and is orthogonal to subsequent parameter-efficient fine-tuning strategies.

In summary, while recent methods have explored global strategies or fine-tuning integration, gaps remain in effectively handling the distinct dynamics of attention heads versus MLP neurons and the critical role of calibration data selection.
Our work addresses these gaps through a theoretically grounded approach that jointly considers optimization dynamics, adaptive sparsity allocation, and calibration-aware pruning.

\subsection{Calibration data.}
Pruning methods often depend on a small \emph{calibration} set to estimate activation or gradient statistics for scoring and pruning decisions.
Recent works~\citep{Williams_2024} highlight that the \emph{selection} of calibration data plays a critical role in the pruning outcome, with factors such as data quality, diversity, and alignment with the model's pretraining distribution shown to significantly influence pruning effectiveness~\citep{bandari2024c4datasetoptimalpruning,ji2024bewarecalibrationdatapruning}.

\subsection{Neural Tangent Kernel (NTK)}
\label{app:ntk}
\textbf{SGD Perspective.}
Consider training via continuous-time gradient descent (GD) with learning rate $\eta$, 
where the parameter vector $\mathbf{W}_t$ evolves over time $t$.
For a neural network $f(\mtx x;\mathbf{W})$ with training loss $\mathcal{L}$, 
one can write\citep{NEURIPS2019_0d1a9651}:
\begin{align*}
\dot{\mathbf{W}}_t 
&=
-\,\eta
\nabla_{\mathbf{W}_t}f(\mtx x;\mathbf{W}_t)^\top 
\nabla_{f(\mtx x;\mathbf{W}_t)} \mathcal{L},
\\[4pt]
\dot{\mathcal{L}}
&=
\nabla_{f(\mtx x;\mathbf{W}_t)} \mathcal{L}^\top 
\nabla_{\mathbf{W}_t}f(\mtx x;\mathbf{W}_t)
\dot{\mathbf{W}}_t\\
&=
-\eta
\nabla_{f} \mathcal{L}^\top
\Bigl[
  \nabla_{\mathbf{W}} f(\mtx x;\mathbf{W}_t)
  \nabla_{\mathbf{W}} f(\mtx x;\mathbf{W}_t)^\top
\Bigr]
\nabla_{f} \mathcal{L}.
\end{align*}
The factor 
$
\nabla_{\mathbf{W}} f(\mtx x;\mathbf{W}_t)\,\nabla_{\mathbf{W}} f(\mtx x;\mathbf{W}_t)^\top
$
is called the \emph{Neural Tangent Kernel (NTK)}\citep{jacot2018neural} under SGD, denoted
\begin{align*}
\widehat{\Theta}^{\mathrm{SGD}}(\mtx x,\mtx x)
=
\nabla_{\mathbf{W}}f(\mtx x;\mathbf{W})
\nabla_{\mathbf{W}}f(\mtx x;\mathbf{W})^\top=
\bigl\langle
  \nabla_{\mathbf{W}}f(\mtx x;\mathbf{W}),
  \nabla_{\mathbf{W}}f(\mtx x;\mathbf{W})
\bigr\rangle.
\end{align*}
\textbf{Adam (SignGD) Perspective.}
Modern Transformer-based language models commonly use \emph{Adam} 
rather than plain SGD. 
Considering Adam's exact analysis is more complicated,
existing work \citep{li2024optimizationgeneralizationtwolayertransformers,zou2021understandinggeneralizationadamlearning,kunstner2023noisemainfactorgap,pmlr-v202-wei23b} 
suggests that \emph{Sign Gradient Descent} (SignGD) often behaves similarly in training dynamics. 
Hence, as a proxy for Adam, we consider a \emph{sign-based} update:
\begin{equation*}
\dot{\mathbf{W}}_t 
=
-\eta\,
\mathrm{sign}
\Bigl(
  \nabla_{\mathbf{W}}f(\mtx x;\mathbf{W}_t)^\top 
  \nabla_{f(\mtx x;\mathbf{W}_t)} \mathcal{L}
\Bigr).
\end{equation*}
By the chain rule,
\begin{align*}
\dot{\mathcal{L}}
&=
\nabla_{f(\mtx x;\mathbf{W}_t)} \mathcal{L}^\top 
\nabla_{\mathbf{W}}f(\mtx x;\mathbf{W}_t)
\dot{\mathbf{W}}_t\\
&=
-\eta
\nabla_{f} \mathcal{L}^\top
\Bigl[
  \nabla_{\mathbf{W}} f(\mtx x;\mathbf{W}_t)\,
  \mathrm{sign}\bigl(\nabla_{\mathbf{W}} f(\mtx x;\mathbf{W}_t)\bigr)^\top
\Bigr]
\nabla_{f} \mathcal{L}.
\end{align*}
Following the NTK viewpoint, we define the \emph{asymmetric SignGD kernel} as
\begin{align*}
\widehat{\Theta}^{\mathrm{A\text{-}Sign}}(\mtx x,\mtx x)
&=
\nabla_{\mathbf{W}}f(\mtx x;\mathbf{W})\,
\mathrm{sign}\bigl(\nabla_{\mathbf{W}}f(\mtx x;\mathbf{W})\bigr)^\top=
\bigl\langle
  \nabla_{\mathbf{W}}f(\mtx x;\mathbf{W}),
  \mathrm{sign}\bigl(\nabla_{\mathbf{W}}f(\mtx x;\mathbf{W})\bigr)
\bigr\rangle.
\end{align*}
For simplicity, we write $\Theta = \widehat{\Theta}^{\mathrm{A\text{-}Sign}}$ in what follows.

\section{Additional Results}
\label{app:experiment}

\subsection{Experimental Settings}
\label{app:experiment_setup_details}

\textbf{Model backbones.}  
We evaluate \alg~on three open-source large language models covering both decoder-only and encoder-decoder architectures: Llama3.1-8B, Llama3.2-3B \citep{dubey2024llama3herdmodels}, Qwen2.5-7B \citep{qwen2025qwen25technicalreport}, and T5-base \citep{raffel2023exploringlimitstransferlearning}.
Our main experiments are conducted on Llama3.1-8B, with additional results for the other models provided in \cref{app:experiment}.

\textbf{Baselines.}  
We compare \alg~with several state-of-the-art structured pruning baselines, including LLM-Pruner \citep{NEURIPS2023_44956951}, SliceGPT \citep{xia2024shearedllamaacceleratinglanguage}, 
, FLAP \citep{flap}, Olica \citep{he2025olica}, and Tyr-the-pruner \citep{li2025tyrtheprunerstructuralpruningllms}.
For LLM-Pruner, we use the \texttt{param\_second} configuration, which we find performs better in our setting.
For FLAP, we use \texttt{AL-AM} as the compressed structure setting. For Tyr-the-pruner, considering the extensive numbers of the official source code (419,4304 tokens), we change it to the same number as our calibration tokens (4,096). Even after this modification, the whole process still takes more than 10 hours to complete. 
All methods are compared at matched target model sizes for fairness. All methods implementation strictly follows the official existing source code.

\textbf{Evaluation tasks and datasets.}
To comprehensively evaluate the effectiveness of our method, we conduct experiments on three types of tasks.
First, we follow \citep{NEURIPS2023_44956951} to evaluate zero-shot perplexity on WikiText2 \citep{merity2016pointer}, PTB \citep{marcus-etal-1993-building}, and Lambada \citep{paperno-etal-2016-lambada}.
Second, we evaluate zero-shot accuracy on a suite of commonsense reasoning benchmarks, including ARC-easy \citep{clark2018thinksolvedquestionanswering}, 
Winogrande \citep{sakaguchi2021winogrande}, and HellaSwag \citep{zellers2019hellaswagmachinereallyfinish}, as well as SVAMP (math word problem) \citep{patel2021nlpmodelsreallyable}.
Additionally, we assess 3-shot Pass@1 performance on MBPP (code generation) \citep{austin2021programsynthesislargelanguage}, all except SVAMP use the \texttt{lm-eval-harness}~\citep{lm-eval} framework.
All experiments are run on NVIDIA A100 GPUs.

\textbf{Calibration data.}  
All methods, except SliceGPT, use 32 samples from BookCorpus with a sequence length of 128 as calibration data to preserve the zero-shot setting.
For SliceGPT, we follow the original setup using 128 samples from WikiText with a sequence length of 2048, as using BookCorpus leads to significant performance degradation.

\textbf{Gamma value.}
The adaptive sparsity ratio, $\gamma$, is derived from a general analytic formula detailed in \cref{app:gamma}. This formula is broadly applicable, but the resulting numerical value of $\gamma$ is inherently dependent on the specific architectural hyperparameters of each model (e.g., hidden size, number of heads, MLP intermediate dimension). While our main experiment on Llama3.1-8B yielded (which was empirically validated in Figure 4), this value is not universal. We provide the corresponding values used for the other models in our study, which were all derived from the same analytic formula by inputting each model's respective architectural parameters.

\begin{table}[H]
    \centering
    \caption{Gamma values that are used in our experiments.}
    \begin{tabular}{lc}
    \toprule
      Model   &  $\gamma$ \\
    \midrule
     Llama3.1-8B&	3.36\\
Llama3.2-3B&	3.02\\
Qwen2.5-7B	&10.11\\
\bottomrule
    \end{tabular}
    
    \label{tab:gamma_value}
\end{table}

\textbf{LoRA fine-tuning details.}
We conduct recovery tuning experiments using LoRA~\citep{lora} on Alpaca~\citep{alpaca} following \citep{NEURIPS2023_44956951}.
To ensure a strictly fair comparison, all baselines are fine-tuned under identical settings: LoRA rank $r=8$, $\alpha=16$, learning rate $3 \times 10^{-4}$, batch size 32, maximum sequence length 2048, and trained for 2 epochs.

\subsection{Additional Generation Results}

To further validate the method on generation/instruction-following tasks, we added GSM8K and IFEval evaluations.
We report GSM8K on Llama3.1-8B and IFEval on Qwen2.5-7B to use a stable dense-model reference for each benchmark (20\% sparisty) in \cref{tab:gsm8k_ifeval_results}.
These results support the same trend as SVAMP and MBPP. NIRVANA better preserves generation-oriented capabilities, not only multiple-choice accuracy. In particular, on GSM8K, NIRVANA substantially outperforms LLM-Pruner, FLAP, and Tyr-the-Pruner after pruning. On IFEval, NIRVANA remains close to the dense model and performs better than LLM-Pruner and FLAP. Note that we used Qwen2.5-7B for IFEval because we observed a poor performance on it with Llama.

\begin{table}[htbp]
\centering
\begin{tabular}{lcc}
\toprule
Method & GSM8K (8-shot) $\uparrow$ & IFEval $\uparrow$ \\
\midrule
Dense            & 48.98 & 38.73 \\
LLM-Pruner       & 4.47  & 27.22 \\
FLAP             & 3.49  & 29.14 \\
Tyr-the-Pruner   & 3.36  & -- \\
NIRVANA & \textbf{18.27} & \textbf{37.05} \\
\bottomrule
\end{tabular}
\caption{Evaluation Results on GSM8K and IFEval}
\label{tab:gsm8k_ifeval_results}
\end{table}

We further conducted a focused MBPP analysis at 20\% sparsity to compare which structures are preserved by NIRVANA, LLM-Pruner, and FLAP in \cref{tab:arch_retention}. The pattern suggests that NIRVANA does not preserve parameters uniformly. Instead, it preserves many intact attention layers while avoiding the strong early/middle MLP pruning observed in FLAP. LLM-Pruner, in contrast, shows a more uniform MLP retention profile across early/middle/late layers and preserves no fully intact attention layer.
We also measured output drift on MBPP prompts using the dense model as reference in \cref{tab:kl_agreement}. The metric is full-vocabulary KL from the dense next-token distribution to the pruned model distribution at the last prompt token. NIRVANA perturbs the dense model's code-prompt next-token behavior the least, which is consistent with its much higher MBPP pass@1. Finally, we re-executed the stored MBPP generations and found that NIRVANA keeps both code validity and task-solving accuracy: it achieves 92.6\% syntax-valid outputs and 23.8\% pass@1, while LLM-Pruner has many more syntax failures and FLAP often produces syntactically valid code that fails the semantic tests.

\begin{table}[htbp]
\centering
\resizebox{\textwidth}{!}{%
\begin{tabular}{lccccccc}
\toprule
Method & MBPP pass@1 & Mean Attn Ret. & Mean MLP Ret. & Full Attn Layers & Early MLP Ret. & Mid MLP Ret. & Late MLP Ret. \\
\midrule
NIRVANA & 23.8 & 93.4\% & 76.6\% & 22 & 94.0\% & 77.0\% & 58.6\% \\
LLM-Pruner & 5.0 & 87.5\% & 80.0\% & 0 & 80.0\% & 80.0\% & 80.0\% \\
FLAP & 2.4 & 100.0\% & 74.8\% & 32 & 66.3\% & 65.0\% & 95.1\% \\
\bottomrule
\end{tabular}%
}
\caption{Architecture Retention and MBPP pass@1}
\label{tab:arch_retention}
\end{table}

\begin{table}[htbp]
\centering
\resizebox{\textwidth}{!}{%
\begin{tabular}{lcccc}
\toprule
Method & KL $\downarrow$ & Median KL $\downarrow$ & Dense Top-1 Agreement $\uparrow$ & Dense Top-1 in Pruned Top-5 $\uparrow$ \\
\midrule
NIRVANA & \textbf{0.1107} & \textbf{0.0298} & \textbf{90.0\%} & \textbf{99.2\%} \\
LLM-Pruner & 0.3779 & 0.1938 & 79.4\% & 97.0\% \\
FLAP & 0.4062 & 0.1500 & 65.0\% & 96.0\% \\
\bottomrule
\end{tabular}%
}
\caption{KL Divergence and Top-K Agreement}
\label{tab:kl_agreement}
\end{table}

\subsection{Additional Results on Llama-3.2-3B}
\label{app:experiment_llama3b}
As shown in \cref{tab:llama_3b}, similar to the main experiment in \cref{tab:llama}, our method consistently achieves the best performance across all sparsity levels and evaluation tasks, particularly demonstrating strong robustness in both perplexity and downstream zero-shot tasks. Compared to LLM-Pruner and SliceGPT, our method yields significantly lower perplexity and higher task accuracy under the same compression ratio, highlighting its effectiveness in preserving both language modeling capacity and generalization abilities after pruning.

\begin{table}[H]
    \centering
    \caption{Evaluation results of Llama3.2-3B. \textbf{Bold} indicates the best results while \underline{underline} indicates the second-best.}
    \resizebox{\linewidth}{!}{
    \begin{tabular}{ll|ccc|cccc|c}
        \toprule
        \toprule
        Sparsity& Method & WikiT $\downarrow$ & PTB $\downarrow$ & LambD $\downarrow$ & ARC-e & WinoG &HellaS& MBPP$^\star$   & \#Param\\
        \midrule        
        \multirow{1}{*}{0\%} & Llama3.2-3B& 10.42 & 16.91 & 22.76 & 74.54 & 69.38 & 73.55& 38.00&3.21B\\
        \midrule
        \midrule
        \multirow{3}{*}{\parbox{1cm}{20\%}} 
        & LLM-Pruner & 24.61& \underline{58.12} & \underline{37.52} & \textbf{59.81} & \underline{58.25} & \underline{54.55}& \underline{3.60}  & 2.70B  \\
        & SliceGPT & \underline{23.85} & 144.41 & 220.20 & 46.59 & 57.85 &44.06& 0.00 & 3.31B  \\
        & \alg & \textbf{17.73} & \textbf{26.20} & \textbf{33.12} & \underline{59.64}& \textbf{59.35}&\textbf{55.40}& \textbf{16.40} & 2.65B  \\
        \midrule
        \multirow{3}{*}{\parbox{1cm}{50\%}} & LLM-Pruner & 252.46& \underline{367.33}& \underline{179.02}& 29.17 & 49.17 &26.70& 0.00  & 1.80B\\
        & SliceGPT & \underline{83.25} & 670.36 &  691.64 & \underline{29.63} & \textbf{50.83} & \underline{28.72} & 0.00 & 2.15B \\
        & \alg~(ours) &  \textbf{79.44} & \textbf{108.58} & \textbf{119.25} & \textbf{34.68}& \underline{50.59}&\textbf{33.46} & 0.00 & 1.81B \\
        \bottomrule
        \bottomrule
    \end{tabular}
    }
    \label{tab:llama_3b}
    \caption*{\raggedright\tiny\textsuperscript{$\quad\star$}Pass@1. 3-shot.}
    \vskip -0.15in
\end{table}

\subsection{Zero-shot on Qwen2.5-7B \& Qwen2.5-14B}
\label{app:experiment_qwen}
\cref{tab:qwen} shows the evaluation results on Qwen2.5-7B, and \cref{tab:qwen_14b} shows the results on Qwen2.5-14B. Across different sparsity levels, our method consistently achieves the best overall performance. Notably, at the extreme sparsity level of 50\%, LLM-Pruner encounters severe degradation and unstable outputs, while our approach still maintains reasonable perplexity and accuracy, demonstrating stronger robustness under high compression ratios.

\begin{table}[H]
    \centering
    \caption{Evaluation results of Qwen2.5-7B. \textbf{Bold} indicates the best results.}
    \resizebox{\linewidth}{!}{
    \begin{tabular}{ll|ccc|cclc|c}
        \toprule
        \toprule
        Sparsity& Method & WikiT $\downarrow$ & PTB $\downarrow$ & LambD $\downarrow$ & ARC-e & WinoG &  HellaS&MBPP$^\star$   & \#Param\\
        \midrule        
        \multirow{1}{*}{0\%} & Qwen2.5-7B & 9.05 & 15.64 & 22.41 & 80.47& 72.85&  78.86&64.20&7.62B\\
        \midrule
        \midrule
        \multirow{2}{*}{\parbox{1cm}{20\%}} & LLM-Pruner & 17.73 & 33.12 & 34.17 & 60.44& 54.93&  60.30&12.4& 6.27B \\
        & \alg~(ours) & {17.73} & \textbf{29.22} & \textbf{32.10} & \textbf{68.98} & \textbf{65.69} & \textbf{69.24}&  \textbf{32.6}& 6.31B \\
        \midrule
        \multirow{2}{*}{\parbox{1cm}{50\%}} & LLM-Pruner & 8625.69 & 11789.92 & 8625.69 & 24.79& 51.38&  25.73&0.00  & 4.35B \\
        & \alg~(ours) & \textbf{77.00} & \textbf{148.41} & \textbf{79.44} & \textbf{38.43}&\textbf{52.80}&  \textbf{38.50}&0.00 & 4.34B \\
        \bottomrule
        \bottomrule
    \end{tabular}
    }
    \label{tab:qwen}
    \caption*{\raggedright\tiny\textsuperscript{$\quad\star$}Pass@1. 3-shot. }
    \vskip -0.15in
\end{table}

\begin{table}[H]
    \centering
    \caption{Evaluation results of Qwen2.5-14B. \textbf{Bold} indicates the best results.}
    \resizebox{\linewidth}{!}{
    \begin{tabular}{ll|ccc|cclc}
        \toprule
        \toprule
        Sparsity& Method & WikiT $\downarrow$ & PTB $\downarrow$ & LambD $\downarrow$ & ARC-e & WinoG &  HellaS&MBPP$^\star$  \\
        \midrule        
        \multirow{1}{*}{0\%} & Qwen2.5-14B & 7.16 & 13.38 & 20.09 & 79.34 & 75.69 & 82.91 & 69.00\\
        \midrule
        \midrule
        \multirow{2}{*}{\parbox{1cm}{20\%}} & LLM-Pruner &12.18 & 21.05 & \textbf{25.39} & 70.41 & 66.46 & \textbf{73.15} & 8.20 \\
        & \alg~(ours) & \textbf{11.99} & \textbf{19.18} & 26.20 & \textbf{73.40} & \textbf{68.19} & 72.03 & \textbf{41.60}  \\
        \midrule
        \multirow{2}{*}{\parbox{1cm}{50\%}} & LLM-Pruner & 3827.63 & 6310.69 & 5231.74 & 26.56 & 48.93 & 26.08 & 0.00 \\
        & \alg~(ours) & \textbf{59.96} & \textbf{84.56} & \textbf{105.24} & \textbf{37.42} & \textbf{52.57} & \textbf{38.85} & \textbf{0.00} \\
        \bottomrule
        \bottomrule
    \end{tabular}
    }
    \label{tab:qwen_14b}
    \caption*{\raggedright\tiny\textsuperscript{$\quad\star$}Pass@1. 3-shot. }
    \vskip -0.15in
\end{table}

\subsection{Additional Results with Olica}
\label{app:exp_olica}

\begin{table}[H]
\centering
\caption{Comparison with the recent structured pruning method Olica \citep{he2025olica} on Llama3.1-8B. NIRVANA consistently outperforms Olica, particularly at higher sparsity ratios where Olica suffers from performance collapse due to the scarcity of KV heads in GQA.}
\label{tab:comparison_olica}
\begin{tabular}{lcccc}
\toprule
\textbf{Model} & \textbf{Sparsity} & \textbf{WikiT} $\downarrow$ & \textbf{PTB} $\downarrow$ & \textbf{LambD} $\downarrow$ \\
\midrule
Llama3.1-8B & 0\% & 8.50 & 14.02 & 20.09 \\
\midrule
Olica & 20\% & 20.28 & 67.95 & 27.53 \\
\textbf{NIRVANA (Ours)} & 20\% & \textbf{13.38} & \textbf{19.77} & \textbf{26.20} \\
\midrule
Olica & 40\% & 1978.00 & 7207.03 & 1307.38 \\
\textbf{NIRVANA (Ours)} & 40\% & \textbf{28.33} & \textbf{38.72} & \textbf{43.19} \\
\bottomrule
\end{tabular}
\end{table}

Additional to the main experiments, we compare NIRVANA to Olica \citep{he2025olica} on Llama3.1-8B. Notably, we observe a performance collapse for Olica at 40\% sparsity. We verified our reproduction settings and attribute this to a potential structural mismatch with Llama3.1's Grouped Query Attention (GQA). Unlike Multi-Head Attention (MHA), GQA has significantly fewer KV heads with lower redundancy. Olica's pruning metric likely over-prunes these critical KV heads at higher sparsity levels. In contrast, NIRVANA's adaptive ratio explicitly balances the head/neuron reduction, successfully preserving these sensitive components.

\subsection{Full Budget Tyr-the-pruner Results}

In our main experiments, we reduce the search cost for Tyr-the-pruner since it is time-consuming. To further understand how the benefits are scaling based on the search cost, we evaluate it using the full cost in \cref{tab:tyr_full}.
The run took 4 iterations and 17h 37m 42s on a single A100 GPU.
This result shows that the reduced-budget Tyr row in the main table may underestimate Tyr-the-Pruner, especially on multiple-choice benchmarks. At the same time, the result also highlights the trade-off. Compared with full-budget Tyr, NIRVANA still achieves better PPL on all three corpora and substantially better generation/reasoning performance on SVAMP and MBPP, while using a one-shot pruning procedure (in under 2s) rather than a 17.6-hour iterative run.

\begin{table}[htbp]
\centering
\resizebox{\textwidth}{!}{%
\begin{tabular}{lccccccccc}
\toprule
Method & WikiT $\downarrow$ & PTB $\downarrow$ & LambD $\downarrow$ & ARC-e & WinoG & HellaS & SVAMP & MBPP & Avg Acc \\
\midrule
Tyr-the-pruner (in \cref{tab:llama}) & 19.77 & 38.11 & 31.11 & 68.03 & 66.46 & 67.27 & 19.67 & 4.00 & 45.09 \\
Tyr-the-pruner (full budget)  & 14.47 & 22.41 & 27.89 & \textbf{75.46} & \textbf{68.59} & \textbf{73.30} & 31.00 & 8.60 & 51.39 \\
NIRVANA            & \textbf{13.38} & \textbf{19.77} & \textbf{26.20} & 68.52 & 67.40 & 66.75 & \textbf{49.00} & \textbf{23.80} & \textbf{55.09} \\
\bottomrule
\end{tabular}%
}
\caption{Full cost performance comparison with Tyr-the-pruner.}
\label{tab:tyr_full}
\end{table}

\subsection{Fine-tuning on T5}
\label{app:experiment_t5}

Due to the high computational cost of fine-tuning and our observation that attention heads retain more critical information than MLP neurons, we conduct experiments on T5-base \citep{raffel2023exploringlimitstransferlearning} by pruning only the MLP layers, while keeping the attention heads intact. All MLP layers are pruned to a global sparsity of 50\%.

We compare our method with several baselines, including classic unstructured foresight pruning methods originally proposed for vision models---SNIP \citep{lee2019snipsingleshotnetworkpruning}, SynFlow \citep{NEURIPS2020_46a4378f}, and NTK-SAP \citep{wang2023ntksapimprovingneuralnetwork}---as well as LLM-specific pruning methods such as Wanda \citep{sun2023wanda} and LLM-Pruner \citep{NEURIPS2023_44956951}. For foresight baselines, we adapt their saliency criteria within our framework to ensure a fair comparison, while for Wanda and LLM-Pruner we directly follow their official implementations, using the parameter-second variant for LLM-Pruner.

We note that the comparison here focuses exclusively on the fine-tuning setting. This is because structured pruning methods without recovery typically suffer from significant zero-shot degradation, making direct comparison with unstructured methods---which often retain better zero-shot performance---unfair. However, after sufficient supervised fine-tuning (SFT), the performance gap between structured and unstructured methods narrows considerably, allowing for a more meaningful comparison under the fine-tuning scenario.
All experiments are conducted on a single V100-SXM2-32GB GPU.

\begin{table}[ht]
\caption{Evaluation results (accuracy) of t5-base on GLUE datasets. \textbf{Bold} indicates the best results while \underline{underline} indicates the second-best.}
\vskip 0.05in
\centering
\begin{tabular}{lcccc}
\toprule
 & MRPC & CoLA & SST2 & MNLI \\
\midrule
t5-base & 91.42 & 83.89 & 94.84 & 86.27 \\
\midrule
Magnitude & 84.80 & 71.81 & 92.43 & 83.74 \\
SNIP & \underline{90.20} & \underline{82.17} & \textbf{94.50} & \underline{85.86}\\
SynFlow & 85.29 & 78.04 & 90.94 & 80.97 \\
NTK-SAP & 88.24 & 81.30 & 93.35 & 85.16 \\
Wanda & \textbf{90.69} & \underline{82.17} & 93.81 & {85.67} \\
LLM-Pruner & \underline{90.20} & 81.20 & 93.58 & 85.37\\
\midrule
\alg\ (ours) & \textbf{90.69} & \textbf{82.45} & \underline{94.27} & \textbf{85.99} \\
\bottomrule
\end{tabular}
\label{tab:t5}
\end{table}

Table \ref{tab:t5} presents the fine-tuning performance of NIRVANA and baseline methods on selected GLUE tasks. We observe the following:
(1) {NIRVANA consistently outperforms the baseline methods.} This empirically proves our theory that using NTK-based pruning criteria leads to better preservation of the model's learning dynamics compared to other baselines. This is especially impressive since \alg is on par and even manages to surpass the performance of those unstructured pruning methods such as Wanda and SNIP. (2) While SynFlow considers gradients with respect to the model's output, its reliance on synthetic inputs leads to suboptimal results. Similarly, NTK-SAP approximates NTK using first-order Taylor expansion in a weight-agnostic way, weakening the performance under the pre-trained language model scenario. 

\subsection{Efficiency Analysis}
\label{app:efficiency}

\begin{table}[t!]
\caption{Post-pruning statistics during inference, tested on 20 rounds of Wikitext (batch size 32).}
\centering
\resizebox{\linewidth}{!}{
\begin{tabular}{llccccc}
\toprule
Model&	\#Params	&FLOPs (G)&	Latency (s/bch) $\downarrow$&	Throughput (tks/s) $\uparrow$	&Peak Memory (GB)\\
\midrule
Llama-3.1-8B-14336	&8.03	&6.98&	0.32&	12800&	17.95 \\
Uniform-11469(1)	&6.73	&5.75&	0.68&	6024	&15.46 \\
Uniform-11470(2)	&6.73	&5.75&	0.40&	10240&	15.46 \\
Uniform-11468(4)	&6.73	&5.75&	0.39&	10503&	15.46 \\
Uniform-11464(8)	&6.73	&5.75&	0.29&	14124&	15.46 \\
Uniform-11472(16)	&6.73	&5.75&	0.29&	14124&	15.46 \\
Uniform-11488(32)	&6.73	&5.75&	0.29&	14124&	15.46 \\
Layer-wise(8s)&	6.73&	5.75&	0.28&	14629&	15.45\\
Attention-only	&6.73&	6.15&	0.27&	15170&	14.70\\
MLP-only&	6.73	&5.68	&0.28	&14629&	15.54 \\
\bottomrule
\end{tabular}
}
\label{tab:efficiency_check}
\end{table}

We provide the inference efficiency analysis on Llama3.1-8B in \cref{sec:efficiency}. Here, we provide additional experiments on T5 about the fine-tuning efficiency. We also provide experiments to support our conclusion that ensuring a dimension of multiples of 8 is critical to the model's latency and throughput. 

\begin{wraptable}{r}{6.8cm}
\vskip -0.15in
\caption{Post-pruning statistics during SFT, tested on one round of SST2 (batch size 16).}
\centering
\scalebox{0.9}{
\begin{tabular}{lccc}
\toprule
     & \# Param  & Time (s) & Mem (GB) \\
\midrule
t5-base  & 223.50M & 927.91 & 6.78 \\
SNIP & 166.87M & 928.63 & 6.79 \\
\alg & 166.87M & 732.80 & 5.48  \\
\bottomrule
\end{tabular}
}
\label{tab:efficiency_t5}
\end{wraptable}
\textbf{Fine-tuning Efficiency.} Table \ref{tab:efficiency_t5} shows the impact of pruning on computational cost on a single V100. NIRVANA reduces memory usage by 19.2\% and achieves a 1.3$\times$ speedup in fine-tuning, demonstrating its practical advantage over dense models. In contrast, although SNIP utilizes a similar saliency score, it applies unstructured pruning, which is not easily exploitable by standard hardware or deep learning libraries for runtime acceleration. As a result, SNIP's fine-tuning time and memory usage remain nearly identical to the dense model.

\begin{wraptable}{l}{5.8cm}
\vskip -0.2in
\caption{Wallclock time of performing the pruning process on different methods on a single A100.}
\centering
\scalebox{1}{
\begin{tabular}{lccc}
\toprule
     & Time \\
\midrule
LLM-Pruner & 40.19s \\
SliceGPT & 275.98s \\
FLAP & 2.58s \\
Olica & 57.91s\\
Tyr-the-Pruner & 3hrs \\
\alg & 1.58s \\
\bottomrule
\end{tabular}
}
\vskip -0.2in
\label{tab:alg_efficiency}
\end{wraptable}
\textbf{Algorithm Efficiency.} Table \ref{tab:efficiency_t5} shows the impact of pruning on computational cost on a single V100. NIRVANA reduces memory usage by 19.2\% and achieves a 1.3$\times$ speedup in fine-tuning, demonstrating its practical advantage over dense models. In contrast, although SNIP utilizes a similar saliency score, it applies unstructured pruning, which is not easily exploitable by standard hardware or deep learning libraries for runtime acceleration. As a result, SNIP's fine-tuning time and memory usage remain nearly identical to the dense model.

\subsection{Generalizability of our KL-based Calibration Data Selection}

To validate the effectiveness of our KL-divergence-based calibration data selection process, we applied the same KL-based calibration selection to two representative baselines, LLM-Pruner and FLAP at 20\% sparsity in \cref{tab:baseline_kl}. These results isolate two effects. First, KL-based calibration selection is transferable and improves the language-modeling perplexity of existing pruners. Second, calibration alone does not explain NIRVANA's full performance: even after applying KL selection, FLAP and LLM-Pruner remain below NIRVANA on reasoning/code tasks at the same sparsity. This suggests that the saliency criterion and allocation strategy remain important beyond data selection.

\begin{table}[htbp]
\centering
\resizebox{\textwidth}{!}{%
\begin{tabular}{lccccccccc}
\toprule
Method & WikiT $\downarrow$ & PTB $\downarrow$ & LambD $\downarrow$ & ARC-e & WinoG & HellaS & SVAMP & MBPP & Avg Acc \\
\midrule
LLM-Pruner      & 17.45 & 27.89 & 28.33 & \textbf{69.57} & 66.06 & 65.89 & 22.33 & 4.40 & 45.65 \\
LLM-Pruner + KL & 14.02 & 23.12 & 25.79 & 68.48 & 61.72 & 65.38 & 16.00 & 8.80 & 44.07 \\
FLAP            & 20.88 & 31.35 & 31.72 & 60.35 & 66.22 & 59.03 & 16.33 & 3.40 & 41.07 \\
FLAP + KL       & 15.49 & 23.62 & 28.55 & 60.61 & 65.51 & 60.21 & 18.67 & 2.60 & 41.52 \\
NIRVANA         & \textbf{13.38} & \textbf{19.77} & \textbf{26.20} & 68.52 & \textbf{67.40} & \textbf{66.75} & \textbf{49.00} & \textbf{23.80} & \textbf{55.09} \\
\bottomrule
\end{tabular}%
}
\caption{Performance Comparison of Compression Methods}
\label{tab:baseline_kl}
\end{table}

\clearpage

\section{Synergy with Quantization}
\label{app:quantization}

This section explores the relationship between our structured pruning method, \alg, and quantization \citep{pmlr-v202-xiao23c,MLSYS2024_42a452cb,frantar2023gptq}, another prominent model compression technique. We demonstrate that they are \textit{orthogonal} and can be combined for compounded efficiency gains.

\subsection{Pruning vs. Quantization} 
Structured pruning and quantization enhance model efficiency through distinct mechanisms:

Structured Pruning modifies the model's architecture by systematically removing entire components like neurons or attention heads. This directly reduces the model's parameter count and, more importantly, the floating-point operations (FLOPs) required for both inference and training.

Quantization operates on the data type level, reducing the numerical precision of weights and/or activations (e.g., from 16-bit floating-point, FP16, to 8-bit integer, INT8). This primarily decreases the model's memory footprint and the memory bandwidth needed during inference.

\subsection{Experimental Results} 
Given that these two methods are orthogonal, they can be effectively combined. To validate this, we conducted experiments applying post-training INT8 quantization using SmoothQuant \citep{pmlr-v202-xiao23c} to our NIRVANA-pruned models. The results in Table 1 confirm that our pruning approach is fully compatible with quantization, enabling users to leverage the benefits of both strategies simultaneously.

\begin{table}[t]
    \centering
    \caption{Performance of Llama3.1-8B with \alg and SmoothQuant INT8 quantization. Lower perplexity (PPL) is better ($\downarrow$). Higher throughput is better ($\uparrow$).}
    \resizebox{\linewidth}{!}{
    \begin{tabular}{l|ccc|ccc}
        \toprule
        \toprule
        \textbf{Method} & \textbf{WikiT $\downarrow$} & \textbf{PTB $\downarrow$} & \textbf{LambD $\downarrow$} & \textbf{Latency (s/batch) $\downarrow$} & \textbf{Throughput (tok/s) $\uparrow$} & \textbf{Peak Memory (GB) $\downarrow$} \\
        \midrule
        Llama3.1-8B (Base) & 8.50 & 14.02 & 20.09 & 0.35 & 11702.86 & 17.95 \\
Llama3.1-8B.int8 & 8.64 & 14.24 & 20.40 & 0.44 & 9309.09 & 11.50 \\ \midrule
NIRVANA-0.2 & 13.38 & 19.77 & 26.20 & 0.49 & 8359.18 & 15.29 \\
NIRVANA-0.2.int8 & 13.59 & 20.09 & 26.60 & 1.15 & 3561.74 & 10.15 \\ \midrule
NIRVANA-0.5 & 48.94 & 70.11 & 70.11 & 0.36 & 11377.78 & 11.27 \\
NIRVANA-0.5.int8 & 49.71 & 70.11 & 70.11 & 0.75 & 5461.33 & 8.08 \\ 
        \bottomrule
        \bottomrule
    \end{tabular}
    }
    \label{tab:quantization}
    \vskip -0.15in
\end{table}

\subsection{Discussion on Combined Efficiency}
A key consideration for both pruning and quantization is their non-linear impact on performance. Pushing either technique to its extreme---very high sparsity for pruning or very low bit-depth for quantization---inevitably leads to a sharp decline in model accuracy. For instance, recent studies show that while 8-bit quantization is relatively stable, moving to 4-bit can cause severe performance degradation for methods like SmoothQuant \citep{Huang2024}.

This suggests that the most practical approach is not to maximize one method alone but to combine them in their respective "sweet spots." Our results support this strategy: moderate pruning with NIRVANA combined with stable INT8 quantization delivers superior all-around efficiency. This avoids the "performance cliff" associated with applying either method too aggressively in isolation.

Furthermore, a critical advantage of structured pruning over post-hoc quantization is its impact on training costs. Quantization is almost exclusively an inference-time optimization. In contrast, pruning creates an architecturally smaller model that is not only faster for inference but is also significantly cheaper to fine-tune or subject to recovery training. This unique ability to reduce the computational cost of the entire training lifecycle is a benefit that quantization alone cannot provide.

\newpage

\section{Pseudocode of \alg}

\cref{alg:pruning_group} presents the pseudocode for \alg.
The algorithm operates on both MLP neurons and attention heads, taking as input the model weights, target overall sparsity \(v\), and a set of calibration data \(\mathcal{D}\).
First, it computes NTK-guided saliency scores for all structured units (i.e., MLP neurons and attention heads) using gradients obtained from the calibration data.
Then, it applies an adaptive sparsity allocation strategy, where the pruning rates for MLP and attention are balanced according to the ratio \(\gamma\).
Finally, pruning is performed via global ranking within each module type, and the corresponding weights are zeroed out.
This procedure ensures that pruning decisions simultaneously consider both model-level sensitivity and module-specific characteristics. Algorithm~\ref{alg:data_selection} outlines the procedure for selecting calibration data using our KL-divergence-based strategy.
Given a full candidate dataset \(\mathcal{D}\), the algorithm samples multiple candidate batches, prunes the model using each batch, and evaluates the KL divergence between the pruned and original model outputs on a fixed evaluation set \(\mathcal{V}\).
The batch that minimizes the average KL divergence is selected as the final calibration set \(\mathcal{C}^*\).
This approach ensures that the selected data leads to minimal output distribution shift after pruning, providing a simple yet effective proxy for calibration data quality.

\begin{algorithm}[thb]
\caption{NIRVANA: NTK-guided Global Structured Pruning with Adaptive Sparsity Allocation}
\begin{algorithmic}[1]
\STATE \textbf{Input:} Model \(f\), weights \(\mathbf{W}\), target sparsity \(v\), pruning data \(\mathcal{D}\), MLP/Attention unit sets \(\mathcal{U}_{\mathrm{MLP}}, \mathcal{U}_{\mathrm{Attn}}\), ratio \(\gamma\)
\STATE \textbf{Output:} Pruned model \(f'\)

\vspace{0.4em}
\STATE \(G \leftarrow \text{compute\_NTK\_gradients}(f, \mathcal{D})\)

\STATE \textbf{/* Compute saliency scores for all units */}
\FORALL{MLP unit \(u \in \mathcal{U}_{\mathrm{MLP}}\)}
    \STATE \(S_{\mathrm{MLP}}(u) \leftarrow \text{aggregate\_saliency}(G, \mathbf{W}(u))\)
\ENDFOR
\FORALL{Attention head \(h \in \mathcal{U}_{\mathrm{Attn}}\)}
    \STATE \(S_{\mathrm{Attn}}(h) \leftarrow \text{aggregate\_saliency}(G, \mathbf{W}(h))\)
\ENDFOR

\vspace{0.4em}
\STATE \textbf{/* Adaptive sparsity allocation using \(\gamma\) */}
\STATE Compute \(v_{\mathrm{Attn}}, v_{\mathrm{MLP}}\) using Eq.(6)

\STATE \textbf{/* Global ranking and pruning */}
\STATE \(\mathcal{U}_{\mathrm{pruned,MLP}} \leftarrow \text{select\_lowest}(S_{\mathrm{MLP}}, v_{\mathrm{MLP}})\)
\STATE \(\mathcal{U}_{\mathrm{pruned,Attn}} \leftarrow \text{select\_lowest}(S_{\mathrm{Attn}}, v_{\mathrm{Attn}})\)

\STATE \textbf{/* Apply pruning masks */}
\STATE Zero out weights for \(\mathcal{U}_{\mathrm{pruned,MLP}}\) and \(\mathcal{U}_{\mathrm{pruned,Attn}}\)

\vspace{0.4em}
\STATE \textbf{return} pruned model \(f'\)
\end{algorithmic}
\label{alg:pruning_group}
\end{algorithm}

\begin{algorithm}[ht]
\caption{KL-based Calibration Data Selection}
\begin{algorithmic}[1]
\STATE \textbf{Input:} 
    Full dataset \(\mathcal{D}\), 
    batch size \(B\), 
    number of trials \(T\),
    model \(f\),
    pruning method \(\text{Prune}(\cdot)\),
    fixed evaluation set \(\mathcal{V}\).

\STATE \textbf{Output:} Best calibration set \(\mathcal{C}^*\).

\vspace{0.4em}
\STATE $\text{KL\_min} \leftarrow +\infty$
\FOR{$t = 1$ {\bfseries to} $T$}
    \STATE $\mathcal{C}_t \leftarrow \text{sample\_batch}(\mathcal{D}, B)$
    \STATE $\hat{f}_t \leftarrow \text{Prune}(f, \mathcal{C}_t)$
    \STATE $\text{KL\_value} \leftarrow 0$
    \FOR{each $\mathbf{x}$ in $\mathcal{V}$}
        \STATE $p \leftarrow f(\mathbf{x})$
        \STATE $q \leftarrow \hat{f}_t(\mathbf{x})$
        \STATE $\text{KL\_value} \mathrel{+}= \sum_{l=1}^{K} p_l \log \frac{p_l}{q_l}$
    \ENDFOR
    \STATE $\text{KL\_value} \leftarrow \text{KL\_value} / |\mathcal{V}|$
    \IF{$\text{KL\_value} < \text{KL\_min}$}
        \STATE $\text{KL\_min} \leftarrow \text{KL\_value}$
        \STATE $\mathcal{C}^* \leftarrow \mathcal{C}_t$
    \ENDIF
\ENDFOR
\STATE \textbf{return} the final calibration dataset $\mathcal{C}^*$
\end{algorithmic}
\label{alg:data_selection}
\end{algorithm}

\newpage

\clearpage
\newpage

\section{Proof of Proposition \ref{prop:ntk}}
\label{app:proof}

\begin{proposition}[Full version of \cref{prop:ntk}]
\label{prop:ntk_proof}

Consider a Transformer model with parameters $W$, and let
$\widetilde{W}$ denote the parameters after structured pruning. Let
$\mathcal{P}$ be the set of pruned structured units, where the
corresponding parameter blocks $\{W_g\}_{g\in\mathcal{P}}$ are disjoint
over the pruned coordinates. For each unit, define
\begin{equation}
S_g
=
\sum_{j\in g}
\left|
\frac{\partial f(X;W)}{\partial W_j}W_j
\right|,
\qquad
\bar{\delta}_g
=
\frac{S_g}{\|W_g\|_1}.
\end{equation}

We make the following local assumptions:

\begin{enumerate}
    \item \textbf{Local smoothness.}
    Along the line segment between $W$ and $\widetilde{W}$,
    \begin{equation}
    \left\|
    \nabla_W^2 f(X;W+t(\widetilde{W}-W))
    \right\|_{\mathrm{op}}
    \leq L_H,
    \qquad
    \forall t\in[0,1].
    \end{equation}

    \item \textbf{Active pruned groups.}
    The effective gradient magnitude of every pruned structured unit is
    lower-bounded:
    \begin{equation}
    \delta_{\mathcal{P}}
    :=
    \min_{g\in\mathcal{P}}\bar{\delta}_g
    >0.
    \end{equation}
    Equivalently,
    $S_g\geq\delta_{\mathcal{P}}\|W_g\|_1$ for every
    $g\in\mathcal{P}$.
\end{enumerate}

If the total saliency of the pruned structured units satisfies
\begin{equation}
\sum_{g\in\mathcal{P}}S_g\leq\epsilon,
\end{equation}
then the perturbation of the same-input Adam/SignGD trace proxy is
bounded by
\begin{equation}
\left|
\widetilde{\Theta}(X,X)-\Theta(X,X)
\right|
\leq
\frac{\sqrt{D}L_H}{\delta_{\mathcal{P}}}\epsilon,
\end{equation}
where $D$ is the number of parameters in the evaluated structured
layers.
\end{proposition}

\begin{proof}
For the same-input scalar proxy considered in this work,
\begin{equation}
\Theta(X,X)
=
\left\langle
\nabla_W f(X;W),
\operatorname{sign}\!\left(\nabla_W f(X;W)\right)
\right\rangle
=
\left\|\nabla_W f(X;W)\right\|_1.
\end{equation}
The corresponding quantity after pruning is
\begin{equation}
\widetilde{\Theta}(X,X)
=
\left\|\nabla_W f(X;\widetilde{W})\right\|_1.
\end{equation}
By the reverse triangle inequality,
\begin{align}
\left|
\widetilde{\Theta}(X,X)-\Theta(X,X)
\right|
&\leq
\left\|
\nabla_W f(X;\widetilde{W})
-
\nabla_W f(X;W)
\right\|_1 \\
&\leq
\sqrt{D}
\left\|
\nabla_W f(X;\widetilde{W})
-
\nabla_W f(X;W)
\right\|_2.
\end{align}

Let $\Delta W=\widetilde{W}-W$. Using the integral form of the
fundamental theorem of calculus,
\begin{equation}
\nabla_W f(X;\widetilde{W})-\nabla_W f(X;W)
=
\int_0^1
\nabla_W^2 f(X;W+t\Delta W)\Delta W
\,dt.
\end{equation}
Therefore, under the local smoothness assumption,
\begin{align}
\left\|
\nabla_W f(X;\widetilde{W})
-
\nabla_W f(X;W)
\right\|_2
&\leq
\int_0^1
\left\|
\nabla_W^2 f(X;W+t\Delta W)
\right\|_{\mathrm{op}}
\|\Delta W\|_2
\,dt \\
&\leq
L_H\|\Delta W\|_2.
\end{align}
It follows that
\begin{equation}
\left|
\widetilde{\Theta}(X,X)-\Theta(X,X)
\right|
\leq
\sqrt{D}L_H\|\Delta W\|_2.
\end{equation}

Because pruning sets the parameters of the units in
$\mathcal{P}$ to zero and the corresponding parameter blocks are
disjoint,
\begin{equation}
\|\Delta W\|_2
\leq
\|\Delta W\|_1
=
\sum_{g\in\mathcal{P}}\|W_g\|_1.
\end{equation}
The active-group condition gives
\begin{equation}
\|W_g\|_1
\leq
\frac{S_g}{\delta_{\mathcal{P}}}
\qquad
\text{for every }g\in\mathcal{P}.
\end{equation}
Consequently,
\begin{equation}
\|\Delta W\|_2
\leq
\frac{1}{\delta_{\mathcal{P}}}
\sum_{g\in\mathcal{P}}S_g
\leq
\frac{\epsilon}{\delta_{\mathcal{P}}}.
\end{equation}
Substituting this result into the preceding inequality yields
\begin{equation}
\left|
\widetilde{\Theta}(X,X)-\Theta(X,X)
\right|
\leq
\frac{\sqrt{D}L_H}{\delta_{\mathcal{P}}}\epsilon.
\end{equation}
\end{proof}

\paragraph{Scope and limitation of the active-group condition.}
The group-level formulation is motivated by the structured nature of
NIRVANA. In our Llama3.1-8B diagnostic, approximately $0.02\%$ of the
individual pruned coordinates have exactly zero output-surrogate
gradients, so a literal coordinate-wise minimum-gradient assumption
does not hold. In contrast, all pruned structured units have nonzero
aggregate saliency. The group-level condition therefore allows isolated
zero-gradient coordinates while requiring each removed unit to retain
a non-degenerate aggregate signal. Nevertheless, if
$\delta_{\mathcal{P}}$ approaches zero, the bound becomes arbitrarily
loose, and if $\delta_{\mathcal{P}}=0$, it provides no nontrivial
guarantee. We therefore use the proposition only as a conditional local
stability argument, not as a tight guarantee for finite-width
pretrained models.

In additional to this theoretical proof, we provide the results of the Centered Kernel Alignment (CKA) \citep{kornblith2019similarityneuralnetworkrepresentations} to empirically support Proposition \ref{prop:ntk}.
The results demonstrate that NIRVANA achieves the highest NTK alignment among all baselines. This provides empirical evidence that our NTK-driven saliency score effectively preserves the NTK structure during pruning, which is consistent with the state-of-the-art recovery fine-tuning performance shown in \cref{tab:llama}.

\begin{table}[H]
    \centering
    \caption{CKA results of the pruned model and the dense model.}
    \begin{tabular}{lc}
    \toprule
      Method   &  Centered Kernel Alignment  \\
    \midrule
      Magnitude   &  0.9658 \\
      LLM-Pruner	&0.9722\\
FLAP	& 0.9924 \\
\textbf{NIRVANA} &	\textbf{0.9972}\\
\bottomrule
    \end{tabular}
    
    \label{tab:CKA}
\end{table}

\newpage

\section{Measuring the Influence of Removing One Attention Head vs. One MLP Neuron}
\label{app:gamma}

To formally quantify the structural importance of different components and determine the optimal sparsity allocation, we analyze a simplified one-layer Transformer model consisting of a single Attention block and an MLP block. We establish the following notation and standard Gaussian initialization assumptions for our derivation in \cref{tab:ph}.

\begin{table}[h]
    \centering
    \caption{Notation and assumptions for the influence derivation.}
    \begin{tabular}{ll}
    \toprule
    \textbf{Symbol} & \textbf{Description} \\
    \midrule
    $L$ & Sequence length (number of tokens) \\
    $d$ & Hidden state dimension (e.g., 4096) \\
    $h$ & Number of attention heads (e.g., 32) \\
    $d_h = d/h$ & Dimension per head (e.g., 128) \\
    $X \in \mathbb{R}^{L \times d}$ & Input activation tensor to the layer \\
    $\sigma_Q^2, \sigma_K^2, \sigma_V^2$ & Variances of the query, key, and value activations \\
    $\sigma_W^2$ & Variance of the linear layer weights \\
    $\sigma_\phi^2$ & Variance of the FFN activations \\
    \bottomrule
    \end{tabular}
    \label{tab:ph}
\end{table}

\subsection{Attention Head Influence}
Without loss of generality, we use Multi-Head Attention (MHA) to derive the expected influence, which can be naturally extended to Grouped Query Attention (GQA). For a given head $i$, let $Q_i = X W_i^Q \in \mathbb{R}^{L \times d_h}$. The attention weights are defined as $A_i = \mathrm{softmax}\bigl(Q_i K_i^\top / \sqrt{d_h}\bigr) \in \mathbb{R}^{L \times L}$, yielding the context matrix $H_i = A_i V_i \in \mathbb{R}^{L \times d_h}$. 

The final output contribution of this single attention head $i$ is $\Delta_i(X) = Y_i = H_i W_i^O \in \mathbb{R}^{L \times d}$. We measure the structural influence of removing this head by the squared Frobenius norm of its output, $I_i^{\rm attn}(X) = \bigl\lVert \Delta_i(X)\bigr\rVert_F^2$. We compute its expectation as follows:

\begin{align*}
\mathbb{E}\bigl[I_i^{\rm attn}(X)\bigr] 
&= \mathbb{E}\bigl[\|H_i W_i^O\|_F^2\bigr] \\
&= \sum_{t=1}^{L} \mathbb{E}\bigl[\|H_i[t] W^O_i\|_2^2\bigr] \quad (\text{summing over token index } t) \\
&= \sum_{t=1}^L \mathbb{E}\left[\sum_{j=1}^d (H_i[t] W_i^O[:,j])^2\right] \\
&\overset{(1)}{=} d \sigma_W^2 \sum_{t=1}^L \mathbb{E}\left[\|H_i[t]\|_2^2\right] \\
&\overset{(2)}{=} s L d d_h \sigma_W^2 \sigma_V^2
\end{align*}

\textbf{Derivation Details:}
\begin{itemize}
    \item \textbf{(1)} Conditioning on $H_i[t]$ and using the law of total expectation, we have $\mathbb{E}\left[(H_i[t] W_i^O[:,j])^2 \mid H_i[t]\right] = \sigma_W^2 \|H_i[t]\|_2^2$. Summing over the $d$ dimensions and removing the conditioning yields the step.
    \item \textbf{(2)} For a token position $t$ with attention weights $\alpha_{ts} = A_i[t,s]$, the context vector is $H_i[t] = \sum_{s=1}^L \alpha_{ts} V_i[s]$. The expected squared norm is:
    \begin{align*}
    \mathbb{E}\left[\|H_i[t]\|_2^2\right] 
    &= d_h \mathbb{E}\bigl[[H_i[t]]_a^2\bigr] \\
    &= d_h \sum_{s=1}^L \alpha_{ts}^2 \underbrace{\mathbb{E}[V_{i}[s]_a^2]}_{\sigma_V^2} \\
    &= d_h \sigma_V^2 \sum_{s=1}^L \alpha_{ts}^2 \\
    &= s d_h \sigma_V^2
    \end{align*}
    Here, $s = \sum_{s=1}^L \alpha_{ts}^2 \in [1/L, 1]$ represents the sum of squared attention weights (the inverse participation ratio), bounded between uniform attention ($1/L$) and one-hot attention ($1$). Summing over $L$ tokens gives the final term $s L d d_h \sigma_W^2 \sigma_V^2$.
\end{itemize}

\subsection{MLP Neuron Influence}
Similarly, we define the output contribution of a single MLP intermediate neuron. Let the perturbation of removing one neuron be:
\[
\Delta Y_{\text{neuron}} = \phi(X W^{\text{in}}) W^{\text{out}}
\]
where $W^{\text{in}}$ and $W^{\text{out}}$ represent the corresponding column and row slices of the input and output weight matrices, respectively. 

Under standard Gaussian initialization assumptions where $W^{\text{in}} \sim \mathcal{N}(0,\sigma_W^2\mathbf{I})$ and $W^{\text{out}} \sim \mathcal{N}(0,\sigma_W^2\mathbf{I})$, the expected squared Frobenius norm of this perturbation is straightforward to compute:
\[
\mathbb{E}\bigl[\|\Delta Y_{\text{neuron}}\|_F^2\bigr] = \mathbb{E}\bigl[\|\phi(X W^{\text{in}}) W^{\text{out}}\|_F^2\bigr] = L d \sigma_\phi^2 \sigma_W^2
\]

\subsection{Deriving the Optimal Pruning Ratio $\gamma$}
We now instantiate our analytic ratio to determine the optimal sparsity allocation between the Attention and MLP layers. Considering that modern LLM architectures typically employ Grouped Query Attention (GQA) with $h_{kv}$ key/value groups, the expected influence ratio between an entire attention head and a single MLP neuron is:
\[
\frac{I_{\text{attn}}}{I_{\text{mlp}}} 
= \frac{\mathbb{E}\bigl[\|\Delta_i(X)\|_F^2\bigr]}{\mathbb{E}\bigl[\|\Delta Y_{\text{neuron}}\|_F^2\bigr]}
= \frac{h_{kv} s L d d_h \sigma_W^2 \sigma_V^2}{L d \sigma_\phi^2 \sigma_W^2}
= \frac{h_{kv} s d_h \sigma_V^2}{\sigma_\phi^2}
\]

To compute the final structural pruning ratio $\gamma$, we must account for the parameter count difference between removing an attention head ($\#_{\text{attn}}$) and an MLP neuron ($\#_{\text{mlp}}$). The adjusted, parameter-normalized pruning ratio $\gamma$ becomes:
\[
\gamma = \frac{I_{\text{attn}} / \#_{\text{attn}}}{I_{\text{mlp}} / \#_{\text{mlp}}} = \frac{I_{\text{attn}} \cdot \#_{\text{mlp}}}{I_{\text{mlp}} \cdot \#_{\text{attn}}}
\]

Plugging in the specific structural hyperparameters and empirical variances for Llama-3.1-8B, we analytically obtain $\gamma \approx 3.36$. As demonstrated in \cref{fig:gamma}, this theoretically derived ratio strongly aligns with empirical grid-search results, justifying its use as the default allocation hyperparameter in our method.

\subsection{Empirical Validation}

\begin{figure}[H]
    \centering
    \includegraphics[width=0.5\linewidth]{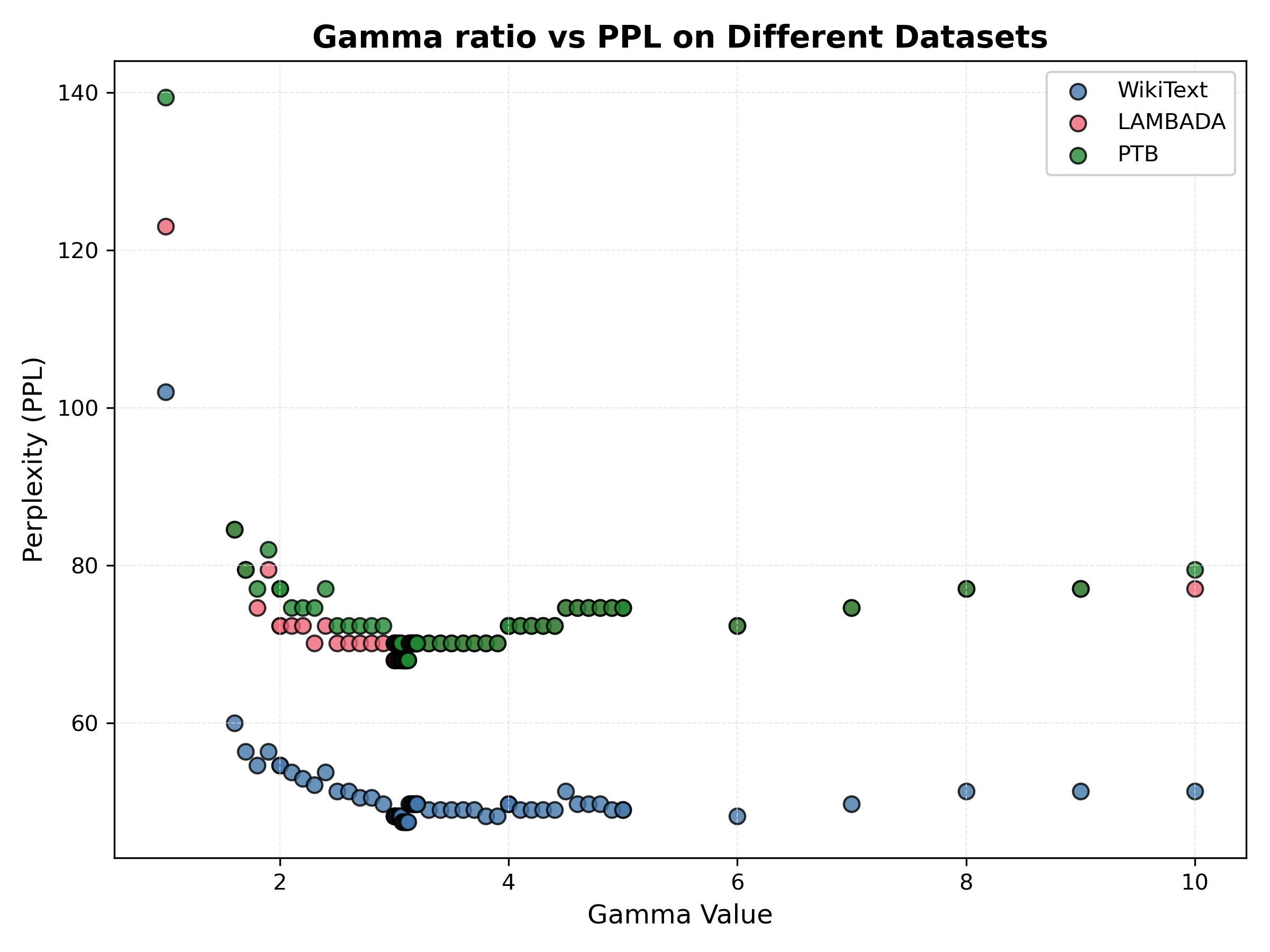}
    \caption{Empirical validation of the sparsity allocation ratio $\gamma$. The optimal value obtained via grid search closely matches our analytically derived ratio, confirming its practical effectiveness.}
    \label{fig:gamma}
\end{figure}

To verify the effectiveness of the pruning ratio \(\gamma\), we conduct empirical experiments with varying \(\gamma\) values, as shown in \cref{fig:gamma}. The best-performing ratio identified through grid search is 3.0, which aligns closely with our analytically derived value of 3.36. This consistency demonstrates that the analytically obtained ratio is well-justified and effective in practice.

We also provide a grid search over Qwen2.5-7B to confirm the gamma value derivation in \cref{tab:gamma_qwen}.

\begin{table}[htbp]
\centering
\begin{tabular}{lccc}
\toprule
Gamma & WikiT $\downarrow$ & PTB $\downarrow$ & LambD $\downarrow$ \\
\midrule
1     & 686.26 & 998.50 & 486.63 \\
3     & 268.74 & 1282.09 & 244.69 \\
5     & 222.79 & 1408.10 & 244.69 \\
10    & 78.42 & \textbf{148.41} & \textbf{78.13} \\
\textbf{10.11} & 77.00 & \textbf{148.41} & 79.44 \\
15    & 102.00 & 196.62 & 126.94 \\
20    & \textbf{76.65} & 156.18 & 78.21 \\
\bottomrule
\end{tabular}
\caption{Grid search of Gamma over Qwen2.5-7B}
\label{tab:gamma_qwen}
\end{table}

\newpage

\section{Formal Justification of the KL-based Calibration Data Selection Process}
\label{app:kl_justification}

\subsection{Theoretical Motivation: Monte Carlo Approximation} 

The goal of pruning can be seen as to find a subnetwork $Q$ that minimizes the expected prediction error relative to the dense teacher $P$ over the true data distribution $\mathcal{D}$: $\min \mathbb{E}_{x \sim \mathcal{D}}[H(P(\cdot|x), Q(\cdot|x))]$.

By the information-theoretic identity $H(P, Q) = H(P) + D_{KL}(P \| Q)$, minimizing the cross-entropy (perplexity) is mathematically equivalent to minimizing the KL divergence. Since the true distribution $\mathcal{D}$ is intractable, we formally treat our anchor set $\mathcal{V}$ as a Monte Carlo estimator \citep{amini2025betterestimationkullbackleiblerdivergence}. The average KL divergence on $\mathcal{V}$ is an unbiased estimator of the true expected divergence:

$$\mathcal{L}_{proxy}(\mathcal{V}) = \frac{1}{|\mathcal{V}|} \sum_{x \in \mathcal{V}} D_{KL}(P(\cdot|x) \| Q(\cdot|x)) \approx \mathbb{E}_{x \sim \mathcal{D}}[D_{KL}(P \| Q)]$$

By selecting the calibration set $\mathcal{C}^*$ that minimizes this proxy, we are statistically maximizing the likelihood that the pruned model behaves like the teacher on the general distribution.

\subsection{Empirical Evidence \& Robustness}

We validate this estimator's effectiveness and stability in two ways:

1. \textbf{Correlation (\cref{fig:kl_data_select})}: We show a strong correlation between our proxy (KL on $\mathcal{V}$) and final downstream performance (PPL), confirming that the estimator is effective.

2. Stability (Robustness Check): To verify that our estimator is not sensitive to the specific sampling of $\mathcal{V}$, we used three distinct, non-overlapping anchor sets that are constructed with different random seeds (V1, V2, V3) to rank 20 candidates.

\begin{table}[H]
    \centering
    \begin{tabular}{lc}
    \toprule
      Validation Set   &  Top-5 rank selected calibration IDs \\
    \midrule
V1 (seed 0)    & 12, 1, 17, 14, 11   \\
 V2 (seed 1234) & 12, 1, 17, 14, 7  \\
 V3 (seed 42)   & 12, 1, 17, 14, 8  \\
\bottomrule
    \end{tabular}
    \caption{Selected calibration datasets ranks with different, not overlapping anchor set $\mathcal{V}$.}
    \label{tab:anchdor}
\end{table}

As shown, the optimal choice (ID=12) was unanimous, and the Top-4 candidates were identical across all three distinct anchor sets. This confirms that a small sample $\mathcal{V}$ provides a high-fidelity estimator with negligible variance relative to the performance gaps, allowing NIRVANA to consistently and robustly identify the optimal candidate within the search space.

\newpage

\section{Impact of Calibration Data}
\label{app:data_selection}

In this section, we investigate the impact of calibration data on pruning effectiveness from three perspectives: sequence length, number of calibration examples, and data quality. 
These factors are often overlooked in existing studies but can have a significant influence on pruning outcomes, particularly for structured pruning where pruning decisions are made globally and depend heavily on the calibration data distribution.

\begin{figure}[H]
\vspace{-0.26in}
    \centering
    \includegraphics[width=0.6\linewidth]{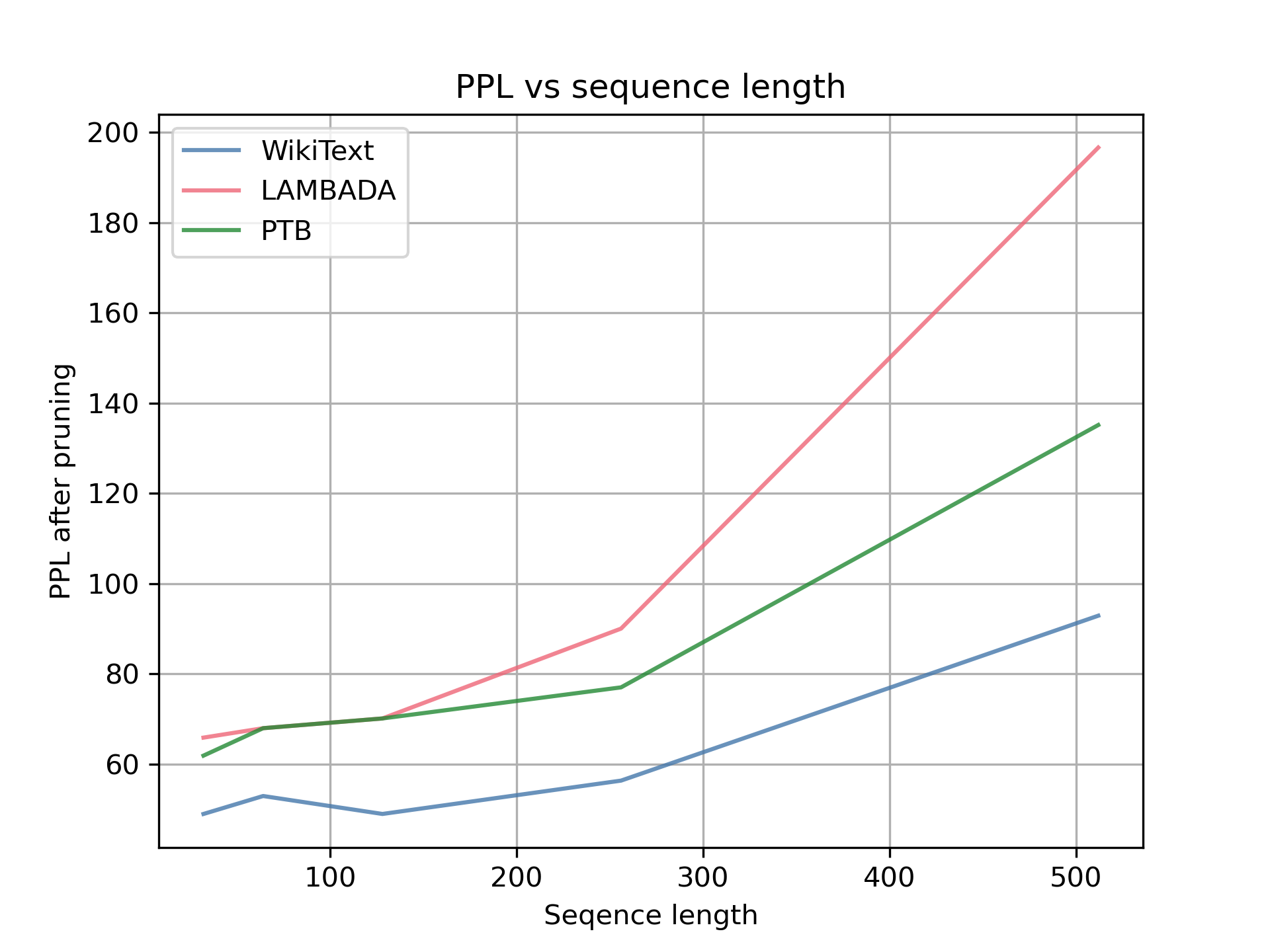}
    \caption{Impact of the sequence length of calibration data. The data are from BookCorpus, with a number of 32.}
    \label{fig:calib_seq_len}
\vspace{-0.2in}
\end{figure}

\textbf{Effect of sequence length.}
We first analyze the impact of sequence length used during calibration.
Longer sequences can provide more representative gradient signals but at the cost of increased computation and memory.
We conduct experiments by varying the sequence length while keeping the number of examples fixed.
\textbf{(Results are reported in \cref{fig:calib_seq_len}.)}

\begin{figure}[H]
\vspace{-0.26in}
    \centering
    \includegraphics[width=0.6\linewidth]{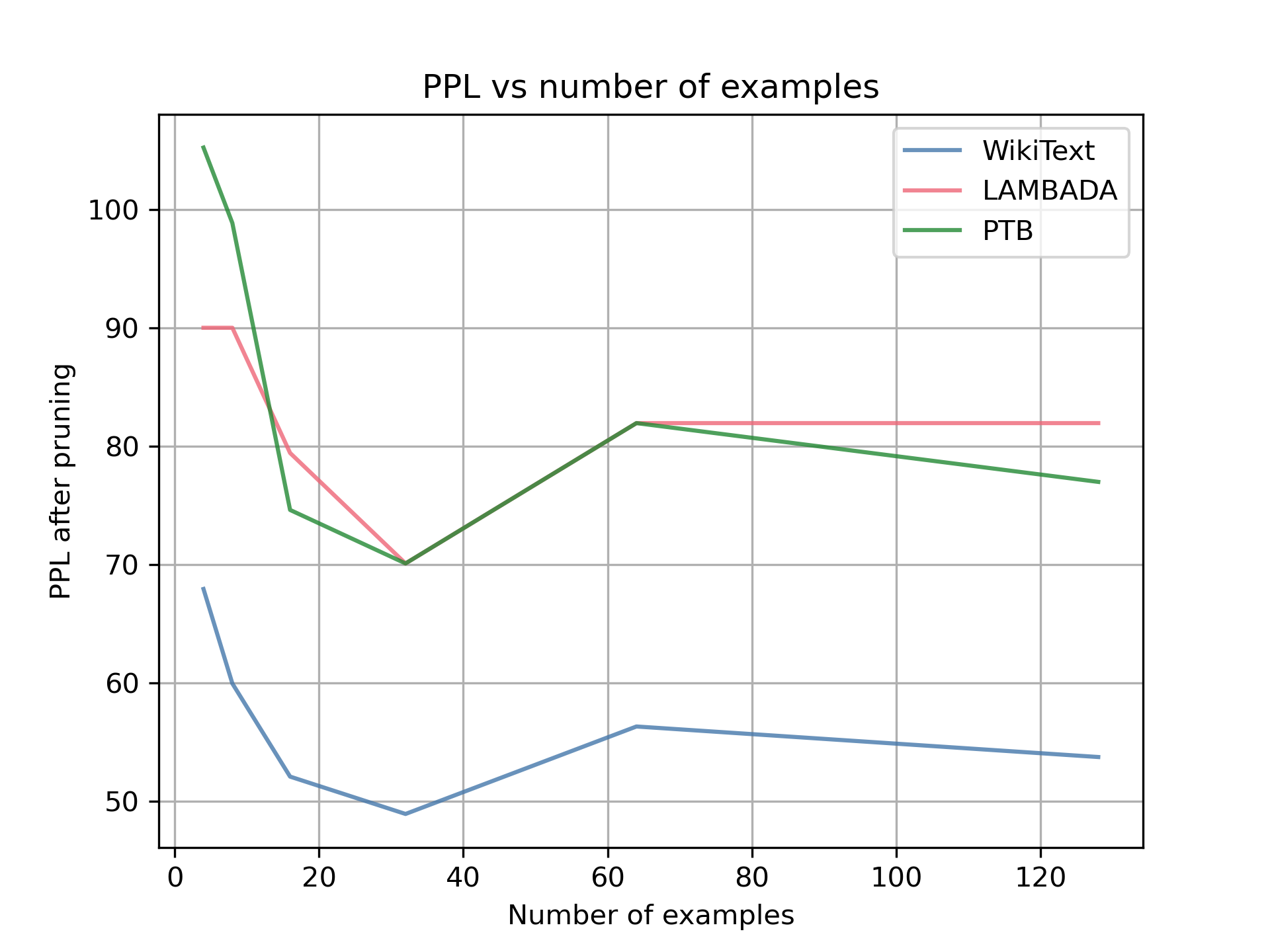}
    \caption{Impact of the number of calibration data examples. The data are from BookCorpus, with a sequence length of 128.}
    \label{fig:calib_num_example}
\vspace{-0.2in}
\end{figure}

\textbf{Effect of number of calibration examples.}
Next, we investigate how varying the number of calibration examples impacts pruning quality while keeping the sequence length fixed.
As shown in \cref{fig:calib_num_example}, we observe that the pruning performance is relatively insensitive to the exact number of examples within the tested range.
Notably, using more calibration data does not consistently improve performance and, in some cases, even leads to degradation.
These results suggest that simply increasing the calibration dataset size does not necessarily provide more effective pruning signals and may introduce noise or redundancy, highlighting that calibration data quality plays a more critical role than quantity.

\textbf{Effect of calibration data quality.}
\cref{tab:calib_source_effect} provides a fine-grained qualitative analysis of the impact of different calibration data samples on the pruned model's performance.
Interestingly, we observe that the first example, which is composed of incoherent fantasy-like text with little semantic consistency, results in the best performance across all evaluation datasets.
In contrast, the second example, which is more structured and factually correct, performs notably worse.
The third example, resembling stream-of-consciousness writing with complex but nonsensical constructs, leads to the worst performance.
These observations suggest that the surface-level quality or coherence of calibration data may not be the primary factor driving pruning effectiveness.
Rather, certain statistical properties or activation patterns---regardless of the semantic content---might play a more dominant role in determining pruning outcomes.
This highlights the importance of further investigating data characteristics beyond human-perceived quality for calibration in pruning.

\begin{table}[t!]
\centering
\caption{Impact of different calibration data samples on the pruned model performance (Perplexity). Lower is better.}
\label{tab:calib_source_effect}
\begin{tabular}{>{\raggedright\arraybackslash}p{7cm}|ccc}
\toprule
\textbf{Calibration Data Sample} & \textbf{WikiText} & \textbf{PTB} & \textbf{Lambada} \\
\midrule
guard at the keep rox - guardians of ambros and yarilo sagi - mate of asok\/seer saratquan - warrior lord of southern city-state sarehl - eldest son of melas and alfar\/strategos sarssen - third ranked warrior - tempkar\/churchik\/yazd sasqua - churchik\/bethel's mate seignore - adept of the conclave setoni - adept of the conclave soji - daughter of elite haskar alleghy\/mate of luton\/leontok strategos - sarehl sushi - large northern duchy sven 
& 105.24 & 157.98 & 209.30 \\
\midrule
fruitless, but more an impelling rallying call for islamic raiders, who gained control of spain and the cherished holy lands in the middle east at the start of the 8th century ; which in 1096 led to the first of nine crusades, resulting in over two hundred years of bloody massacres, merciless victories, and brutal defeats, with power shifting struggles ultimately ending in abject failure for both military campaigns and in a moral hypocrisy that forever stained the offending faiths ; but the power struggles did not end in faraway lands as religious abuses, internal conflicts, and territorial disputes still reigned supreme in the hom    & 222.79 & 471.66 & 295.15 \\
 \midrule
ever be able to travel faster than ten kilometres per haca sure as no tree will ever be able to grow more than ten items of fruit per year, eleven being an unlucky number and thus perpetually avoided in nature ; sure as protein can only ever come from the remains of slaughtered animals ; and sure as there will never be any evidence in favour of life outside glix before pushing his foot down upon the spike the pedal having already been invented, but naturally rejected in favour of the electroconductive abilities of the spike and hurtling forward into a brick wall at 57.11kmph and being crushed between his concrete chair ( concrete being  & 534.46 & 938.00 & 486.63 \\
\bottomrule
\end{tabular}
\end{table}

\clearpage
\newpage

\section{Generation Samples from Pruned Models}
\label{app:sample}

Here we present representative generation samples from pruned models under different sparsity settings for quantitavely and  qualitative comparison. At moderate sparsity levels (20\%), most methods, including \alg, still generate generally coherent and context-relevant responses, though some factual inaccuracies or repetitions can be seen in baselines such as LLM-Pruner and FLAP.
As sparsity increases to 40\% and 50\%, the degradation becomes more pronounced, especially for baseline methods. Common issues include excessive repetition, off-topic content, and hallucinated facts (e.g., "Mount Vesuvius" in FLAP).
Notably, \alg\ tends to preserve better factual grounding and sentence fluency across sparsity levels, although at extreme sparsity (50\%), even \alg\ shows degraded generation quality, with shorter and more generic outputs.
Recovery fine-tuning significantly mitigates these effects for all methods, bringing the generations closer to the original model in both fluency and factuality. However, residual errors remain, and hallucinations can still occur, particularly in models with higher initial sparsity.

\subsection{Quantitatively Analysis of the Generated Examples}

We first provide the quantitative calculated ROUGE \citep{lin-2004-rouge} and BERTScore \citep{zhang2020bertscoreevaluatingtextgeneration} for all generated samples.

\begin{table}[H]
    \centering
    \begin{tabular}{lcc}
    \toprule
      Model   &  ROUGE & BERTScore \\
    \midrule
LLM-Pruner&	0.2858	&0.8586 \\
FLAP&	0.2672&	0.8667\\
NIRVANA (Ours)&	\textbf{0.3035}	& \textbf{0.8732}\\
\bottomrule
    \end{tabular}
    \caption{Quantitative analysis of the generated examples using ROUGE and BERTScore.}
    \label{tab:quant_rslt}
\end{table}

NIRVANA achieves both the highest ROUGE and BERTScore, demonstrating superior semantic alignment and content recall with the ground truth.

\subsection{Quanlitatively Analysis of the Generated Examples}

Furthermore, we adopted a qualitative standard human evaluation framework following established protocols in text generation evaluation \citep{bhandari-etal-2020-evaluating,10.1145/3571730}:
\begin{itemize}
    \item Fluency (F): The sentence-level grammaticality and naturalness of the generated text (local linguistic quality).
\item Relevance (R): Whether the output directly addresses the question in the prompt and stays on topic.
\item Coherence (C): Paragraph-level logical consistency and structural connectedness of the response.
\item Hallucination (H): Whether the output contains fabricated or semantically inconsistent information that is not grounded in the dense model's own distribution, such as inventing nonexistent entities, contradictory statements, or implausible claims.
\end{itemize}

Using this framework, we show that NIRVANA demonstrates superior factual consistency and significantly lower hallucination rates in practical regimes. We provide a full representative example at 20\% sparsity from the fine-tuned model below.

\newpage

\begin{table}[H]
\centering
\caption{Generation samples of recovery fine-tuned pruned models based on Llama3.1-8B at 20\% sparsity.
We also report qualitative ratings for Fluency (F), Relevance (R), Coherence (C), and Hallucination severity (H; lower is better), using $\checkmark$ / $\triangle$ / $\times$ and \textit{low/mid/high}.}
\begin{tabular}{p{3.0cm}|p{10.cm}}
\toprule
\textbf{Prompt} & \textbf{Generated Text} \\
\midrule
\multirow{5}{=}{Who lives in the Imperial Palace in Tokyo?} 
& \textbf{Original model:} The Imperial Palace in Tokyo is the primary residence of the Emperor of Japan. The palace is located in the Chiyoda District of Tokyo. The current Emperor, Akihito, and his family reside in the palace. The palace is open to the public for tours, but only on certain days of the week.\qscore{$\checkmark$}{$\checkmark$}{$\checkmark$}{N} \\
\cmidrule{2-2}
& \textbf{LLM-Pruner:} The Emperor of Japan. The Emperor of Japan is the head of state and the sovereign of Japan. He is the symbol of the nation and is the head of the Imperial House of Japan. The Emperor lives in the Imperial Palace in Tokyo, which is the residence of the Emperor and his family. The Imperial Palace is located in Chiyoda, Tokyo and covers an area of 43 hectares.\qscore{$\checkmark$}{$\checkmark$}{$\checkmark$}{low} \\
\cmidrule{2-2}
& \textbf{FLAP:} The Imperial Palace is located in Chiyoda-ku, Tokyo, and is the primary residence of the Emperor of Japan, also known as the Emperor of the Empire of Japan. It is surrounded by a large moat, and is protected by an outer wall.\qscore{$\checkmark$}{$\checkmark$}{$\checkmark$}{mid} \\
\cmidrule{2-2}
& \textbf{\alg~(Ours):} The Emperor of Japan lives in the Imperial Palace in Tokyo. The Emperor is the head of the Imperial Family and is the symbol of the nation. The Emperor is a figurehead and has no powers, but is the highest ceremonial head of state.\qscore{$\checkmark$}{$\checkmark$}{$\checkmark$}{No} \\
\midrule
\multirow{5}{=}{Where did Yellowstone National Park get its name?} 
& \textbf{Original model:} Yellowstone is named after the Yellowstone River, which in turn was named by John Colter, a member of the Lewis and Clark Expedition. Colter was the first white man to see the area. He called it the ``Yellow Stone River'' because of the yellowish color of the rock in the river.
Yellowstone is the oldest national park in the world and the first national park in the United States.\qscore{$\checkmark$}{$\checkmark$}{$\checkmark$}{N} \\
\cmidrule{2-2}
& \textbf{LLM-Pruner:} The answer is from the French fur trader, Robert Cenudin, who in 1805 explored the region and described it as a ``champagne fountain.'' The name was adopted by Thomas Jefferson, who named the park after him. The park was created in 1872 as the world's first national park, and it has been a popular tourist destination ever since.\qscore{$\checkmark$}{$\checkmark$}{$\checkmark$}{high} \\
\cmidrule{2-2}
& \textbf{FLAP:} What is the most famous volcano in the park?
Yellowstone National Park is named after the famous explorer, artist, and naturalist John C. Calhoun. He was the first person to visit Yellowstone and was the namesake of the park. The most famous volcano in the park is Mount Vesuvius, also known as ``The Lava Mountain.''\qscore{$\checkmark$}{$\triangle$}{$\times$}{mid} \\
\cmidrule{2-2}
& \textbf{\alg~(Ours):} Yellowstone National Park is named after the name of the river that passes through it. The river is named after the Native American tribe, the Sioux, who called it ``Yellowstone''.\qscore{$\checkmark$}{$\checkmark$}{$\checkmark$}{N} \\
\bottomrule
\end{tabular}
\end{table}

\begin{table}[H]
\centering
\caption{Generation samples of pruned models based on Llama3.1-8B. Here the spasity raio is 20\%.}
\begin{tabular}{p{3.0cm}|p{10.cm}}
\toprule
\textbf{Prompt} & \textbf{Generated Text} \\
\midrule
\multirow{5}{=}{Who lives in the Imperial Palace in Tokyo?} 
& \textbf{Original model:} The Imperial Palace in Tokyo is the primary residence of the Emperor of Japan. The palace is located in the Chiyoda District of Tokyo. The current Emperor, Akihito, and his family reside in the palace. The palace is open to the public for tours, but only on certain days of the week. \qscore{$\checkmark$}{$\checkmark$}{$\checkmark$}{N}\\
\cmidrule{2-2}
& \textbf{LLM-Pruner:} The Imperial Family of Japan is composed of His Majesty Emperor Akihito and Her Majesty Empress Michiko, His Imperial Majesty Emperor Emeritus Showa and Her Imperial Majesty Empress Emerita Michiko... \qscore{$\triangle$}{$\times$}{$\times$}{Mid}\\
\cmidrule{2-2}
& \textbf{FLAP:} The Imperial Palace in Tokyo is a place where the Japanese Imperial Family lives. This is where the Japanese Imperial Family lives. The Japanese Imperial Family lives in the Imperial Palace in Tokyo. This is where the Japanese Imperial Family lives. \qscore{$\checkmark$}{$\checkmark$}{$\times$}{N}\\
\cmidrule{2-2}
& \textbf{\alg~(Ours):} Who is the Emperor of Japan?
The Emperor of Japan is the head of the Imperial family and the longest reigning monarch in world history. He lives in the Imperial Palace in Tokyo and is the only living monarch in the world. \qscore{$\checkmark$}{$\triangle$}{$\checkmark$}{Low}\\
\midrule
\multirow{5}{=}{Where did Yellowstone National Park get its name?} 
& \textbf{Original model:} Yellowstone is named after the Yellowstone River, which in turn was named by John Colter, a member of the Lewis and Clark Expedition. Colter was the first white man to see the area. He called it the "Yellow Stone River" because of the yellowish color of the rock in the river.
Yellowstone is the oldest national park in the world and the first national park in the United States. \qscore{$\checkmark$}{$\checkmark$}{$\checkmark$}{N}\\
\cmidrule{2-2}
& \textbf{LLM-Pruner:}  What is the name of the park? What is the name of the park?... \qscore{$\times$}{$\times$}{$\times$}{N.A.}\\
\cmidrule{2-2}
& \textbf{FLAP:} Yellowstone National Park was named after John M. W. Dawson, who was the first person to discover it. Yellowstone National Park is the largest national park in the United States and is home to more than 2 million people. \qscore{$\times$}{$\checkmark$}{$\checkmark$}{High} \\
\cmidrule{2-2}
& \textbf{\alg~(Ours):} Yellowstone National Park is located in Wyoming, United States. It was named Yellowstone because of the sulfuric color of the water. \qscore{$\checkmark$}{$\checkmark$}{$\checkmark$}{N}\\
\bottomrule

\end{tabular}
\end{table}

\begin{table}[H]
\centering
\caption{Generation samples of pruned models based on Llama3.1-8B. Here the spasity raio is 40\%.}
\begin{tabular}{p{3.0cm}|p{10.cm}}
\toprule
\textbf{Prompt} & \textbf{Generated Text} \\
\midrule
\multirow{5}{=}{Who lives in the Imperial Palace in Tokyo?} 
& \textbf{Original model:} The Imperial Palace in Tokyo is the primary residence of the Emperor of Japan. The palace is located in the Chiyoda District of Tokyo. The current Emperor, Akihito, and his family reside in the palace. The palace is open to the public for tours, but only on certain days of the week. \qscore{$\checkmark$}{$\checkmark$}{$\checkmark$}{N}\\
\cmidrule{2-2}
& \textbf{LLM-Pruner:} The Emperor of the Imperial Palace in Tokyo is a great emperor, and the Emperor of the Emperor in Tokyo is a great emperor, and the Emperor of the Emperor in Tokyo is a great emperor, ... \qscore{$\times$}{$\times$}{$\times$}{N.A.}\\
\cmidrule{2-2}
& \textbf{FLAP:} Who lives in the Imperial Palace in Tokyo? I'm looking for a person who is a little bit younger than me. I'm not sure if it's a man or a woman. I'm looking for a person who is a little bit younger than me. I'm not sure if it's a man or a woman. \qscore{$\times$}{$\times$}{$\times$}{High}\\
\cmidrule{2-2}
& \textbf{\alg~(Ours):} The 4th Prince of the Imperial family, Prince Akitaka, is the only one who has the right to live in the Palace.
Akitaka, the 4th Prince of the Imperial family, is the only one who has the right to live in the Palace. \qscore{$\triangle$}{$\checkmark$}{$\times$}{Mid}\\
\midrule
\multirow{5}{=}{Where did Yellowstone National Park get its name?} 
& \textbf{Original model:} Yellowstone is named after the Yellowstone River, which in turn was named by John Colter, a member of the Lewis and Clark Expedition. Colter was the first white man to see the area. He called it the "Yellow Stone River" because of the yellowish color of the rock in the river.
Yellowstone is the oldest national park in the world and the first national park in the United States. \qscore{$\checkmark$}{$\checkmark$}{$\checkmark$}{N}\\
\cmidrule{2-2}
& \textbf{LLM-Pruner:}  What did it do with Yellowstone National Park? What did it do with Yellowstone National Park? What did it do with Yellowstone National Park?
Yellowstone National Park is a national park in the United States of America. \qscore{$\times$}{$\times$}{$\times$}{N.A.}\\
\cmidrule{2-2}
& \textbf{FLAP:} The National Park was named after Yellowstone National Park. Yellowstone National Park is a famous city of Yellowstone. Yellowstone is the capital city of Yellowstone. \qscore{$\times$}{$\triangle$}{$\times$}{High}\\
\cmidrule{2-2}
& \textbf{\alg~(Ours):} It was the first national park in the United States, and the name was given to the Yellowstone National Park. \qscore{$\checkmark$}{$\times$}{$\checkmark$}{N}\\
\bottomrule

\end{tabular}
\end{table}

\begin{table}[H]
\centering
\caption{Generation samples of recovery fine-tuned pruned models based on Llama3.1-8B. Here the spasity raio is 40\%.}
\begin{tabular}{p{3.0cm}|p{10.cm}}
\toprule
\textbf{Prompt} & \textbf{Generated Text} \\
\midrule
\multirow{5}{=}{Who lives in the Imperial Palace in Tokyo?} 
& \textbf{Original model:} The Imperial Palace in Tokyo is the primary residence of the Emperor of Japan. The palace is located in the Chiyoda District of Tokyo. The current Emperor, Akihito, and his family reside in the palace. The palace is open to the public for tours, but only on certain days of the week. \qscore{$\checkmark$}{$\checkmark$}{$\checkmark$}{N}\\
\cmidrule{2-2}
& \textbf{LLM-Pruner:} The Emperor of Japan
The Emperor of Japan is the head of the Imperial Family and is the ceremonial leader of the nation, and the monarch of Japan. The Emperor of Japan is the head of the Imperial Family and is the ceremonial leader of the nation, and the monarch of Japan. The Emperor of Japan is the head of the Imperial Family and is the ceremonial leader of the nation, and the monarch of Japan. \qscore{$\times$}{$\triangle$}{$\times$}{N.A.}\\
\cmidrule{2-2}
& \textbf{FLAP:} - Who lives in the Imperial Palace in Tokyo?
The Imperial Palace, also known as the White Palace, is the official residence of the Emperor of Japan, the Emperor of Japan being the absolute monarch of Japan. The palace was built in 1914 and is located in the heart of Tokyo. It is a magnificent structure that spans the river and is surrounded by lush greenery and gardens. The palace is a symbol of the Emperor's power and is a place of grandeur and grandeur. \qscore{$\times$}{$\checkmark$}{$\checkmark$}{High}\\
\cmidrule{2-2}
& \textbf{\alg~(Ours):} The Imperial Palace, also known as the Tokyo Palace, is the residence of the Emperor of Japan and the First Empress. The palace is located in the city of Tokyo, in the state of Tokyo, and it is a public property that is owned by the Japanese government. The Emperor and the First Empress live in the palace, and they are the head of the Imperial family and the monarchy of Japan. \qscore{$\triangle$}{$\checkmark$}{$\checkmark$}{Low}\\
\midrule
\multirow{5}{=}{Where did Yellowstone National Park get its name?} 
& \textbf{Original model:} Yellowstone is named after the Yellowstone River, which in turn was named by John Colter, a member of the Lewis and Clark Expedition. Colter was the first white man to see the area. He called it the "Yellow Stone River" because of the yellowish color of the rock in the river.
Yellowstone is the oldest national park in the world and the first national park in the United States. \qscore{$\checkmark$}{$\checkmark$}{$\checkmark$}{N}\\
\cmidrule{2-2}
& \textbf{LLM-Pruner:}  Answer
Yes, Yellowstone National Park is named after the Yellowstone River, which flows through the park. \qscore{$\times$}{$\times$}{$\times$}{High}\\
\cmidrule{2-2}
& \textbf{FLAP:} The park is named after Yellowstone Lake, which was named after the town of the same name in 1852. Yellowstone National Park is also named after the town of the same name, which was founded in 1856. \qscore{$\triangle$}{$\checkmark$}{$\checkmark$}{High}\\
\cmidrule{2-2}
& \textbf{\alg~(Ours):} The name ``Yellowstone'' was first given to the national park in 1828 by the French explorer Pierre Bakhayes. The name was derived from the color of the hot springs in the region, which were a deep and vibrant yellow. The name was first given to the national park in 1828 by the French explorer Pierre Bakhayes. The name was derived from the color of the hot springs in the region, which were a deep and vibrant yellow. \qscore{$\checkmark$}{$\checkmark$}{$\checkmark$}{Mid}\\
\bottomrule

\end{tabular}
\end{table}

\begin{table}[H]
\centering
\caption{Generation samples of pruned models based on Llama3.1-8B. Here the spasity raio is 50\%.}
\begin{tabular}{p{3.0cm}|p{10.cm}}
\toprule
\textbf{Prompt} & \textbf{Generated Text} \\
\midrule
\multirow{5}{=}{Who lives in the Imperial Palace in Tokyo?} 
& \textbf{Original model:} The Imperial Palace in Tokyo is the primary residence of the Emperor of Japan. The palace is located in the Chiyoda District of Tokyo. The current Emperor, Akihito, and his family reside in the palace. The palace is open to the public for tours, but only on certain days of the week. \qscore{$\checkmark$}{$\checkmark$}{$\checkmark$}{N}\\
\cmidrule{2-2}
& \textbf{LLM-Pruner:} The following is the answer to the answer.
The following is the answer to the answer.... \qscore{$\times$}{$\times$}{$\times$}{N.A.}\\
\cmidrule{2-2}
& \textbf{FLAP:} Answers Forum
I'm going to ask you to make me a favor. I would greatly appreciate your help. I would like to know if you would be able to answer the following.
I have been searching for years. I have been searching for many years without any clues. I have been searching for many years without any clues. I have been searching for many years without any clues. I have been searching for many years without any clues. I have been searching for many years without any clues. \qscore{$\times$}{$\times$}{$\times$}{N.A.}\\
\cmidrule{2-2}
& \textbf{\alg~(Ours):} The Imperial Palace is a big, fat, fat, fat, fat, fat, fat, fat, fat, fat, fat, fat, fat, fat, fat \qscore{$\times$}{$\times$}{$\times$}{N.A.}\\
\midrule
\multirow{5}{=}{Where did Yellowstone National Park get its name?} 
& \textbf{Original model:} Yellowstone is named after the Yellowstone River, which in turn was named by John Colter, a member of the Lewis and Clark Expedition. Colter was the first white man to see the area. He called it the "Yellow Stone River" because of the yellowish color of the rock in the river.
Yellowstone is the oldest national park in the world and the first national park in the United States.\\
\cmidrule{2-2}
& \textbf{LLM-Pruner:}  The name of the town?. What is the name of the town? The name of the town? \qscore{$\times$}{$\times$}{$\times$}{N.A.}\\
\cmidrule{2-2}
& \textbf{FLAP:} (Question)
This is a question. I am looking for a name for the Yellowstone National Park, where did it get its name from? I would really appreciate the answer. Thank you, \qscore{$\times$}{$\times$}{$\times$}{N.A.}\\
\cmidrule{2-2}
& \textbf{\alg~(Ours):} What is the name of the 3rd largest city in the world? ...\qscore{$\times$}{$\times$}{$\times$}{N.A.}\\
\bottomrule

\end{tabular}
\end{table}

\begin{table}[H]
\centering
\caption{Generation samples of recovery fine-tuned pruned models based on Llama3.1-8B. Here the spasity raio is 50\%.}
\begin{tabular}{p{3.0cm}|p{10.cm}}
\toprule
\textbf{Prompt} & \textbf{Generated Text} \\
\midrule
\multirow{5}{=}{Who lives in the Imperial Palace in Tokyo?} 
& \textbf{Original model:} The Imperial Palace in Tokyo is the primary residence of the Emperor of Japan. The palace is located in the Chiyoda District of Tokyo. The current Emperor, Akihito, and his family reside in the palace. The palace is open to the public for tours, but only on certain days of the week. \qscore{$\checkmark$}{$\checkmark$}{$\checkmark$}{N} \\
\cmidrule{2-2}
& \textbf{LLM-Pruner:} Tokyo is the capital of the Japanese Empire, and the Imperial Palace is located in the center of the city. \qscore{$\times$}{$\times$}{$\checkmark$}{Mid}\\
\cmidrule{2-2}
& \textbf{FLAP:} Who lives in the Imperial Palace in Tokyo? Who lives in the Imperial Palace in Tokyo? Who lives in the Imperial Palace in Tokyo? Who lives in the Imperial Palace in Tokyo? ... \qscore{$\times$}{$\times$}{$\times$}{N.A.}\\
\cmidrule{2-2}
& \textbf{\alg~(Ours):} Tokyo is the capital of Japan, and the Imperial Palace, known as the Palace of the Emperor, is located in the central city of Tokyo. The palace is a complex of buildings and gardens that serve as the residence and home of the Emperor of Japan. \qscore{$\checkmark$}{$\checkmark$}{$\checkmark$}{N}\\
\midrule
\multirow{5}{=}{Where did Yellowstone National Park get its name?} 
& \textbf{Original model:} Yellowstone is named after the Yellowstone River, which in turn was named by John Colter, a member of the Lewis and Clark Expedition. Colter was the first white man to see the area. He called it the "Yellow Stone River" because of the yellowish color of the rock in the river.
Yellowstone is the oldest national park in the world and the first national park in the United States.  \qscore{$\checkmark$}{$\checkmark$}{$\checkmark$}{N}\\
\cmidrule{2-2}
& \textbf{LLM-Pruner:} Yellowstone National Park was named after the Yellowstone River, which was named after its namesake, the Yellowstone River. The park was created in 1869 by the U.S. Department of Interior, and was initially called the Yellowstone Park. However, the name eventually changed to Yellowstone National Park, and is now a national park.  \qscore{$\checkmark$}{$\checkmark$}{$\checkmark$}{High}\\
\cmidrule{2-2}
& \textbf{FLAP:} Yellowstone National Park (USA)
The Yellowstone National Park was named after the volcano of the same name, which erupted in 1861. The park was created in 1861 by the United States government, which had previously owned the land and wanted to protect it from mining and logging.  \qscore{$\checkmark$}{$\checkmark$}{$\checkmark$}{High}\\
\cmidrule{2-2}
& \textbf{\alg~(Ours):} Yellowstone National Park is named after the name of the river that passes through it. The river is named after the Native American tribe, the Sioux, who called it "Yellowstone".
Where was the first visit to Yellowstone National Park? The first official visit to Yellowstone National Park took place on August 5, 1869. The park was named after the Sioux tribe of North America, who called it "Yellowstone."  \qscore{$\checkmark$}{$\checkmark$}{$\triangle$}{Low}\\
\bottomrule

\end{tabular}
\end{table}

\newpage

\section{Pruning Patterns across Layers}

\begin{figure}
    \centering
    \includegraphics[width=0.8\linewidth]{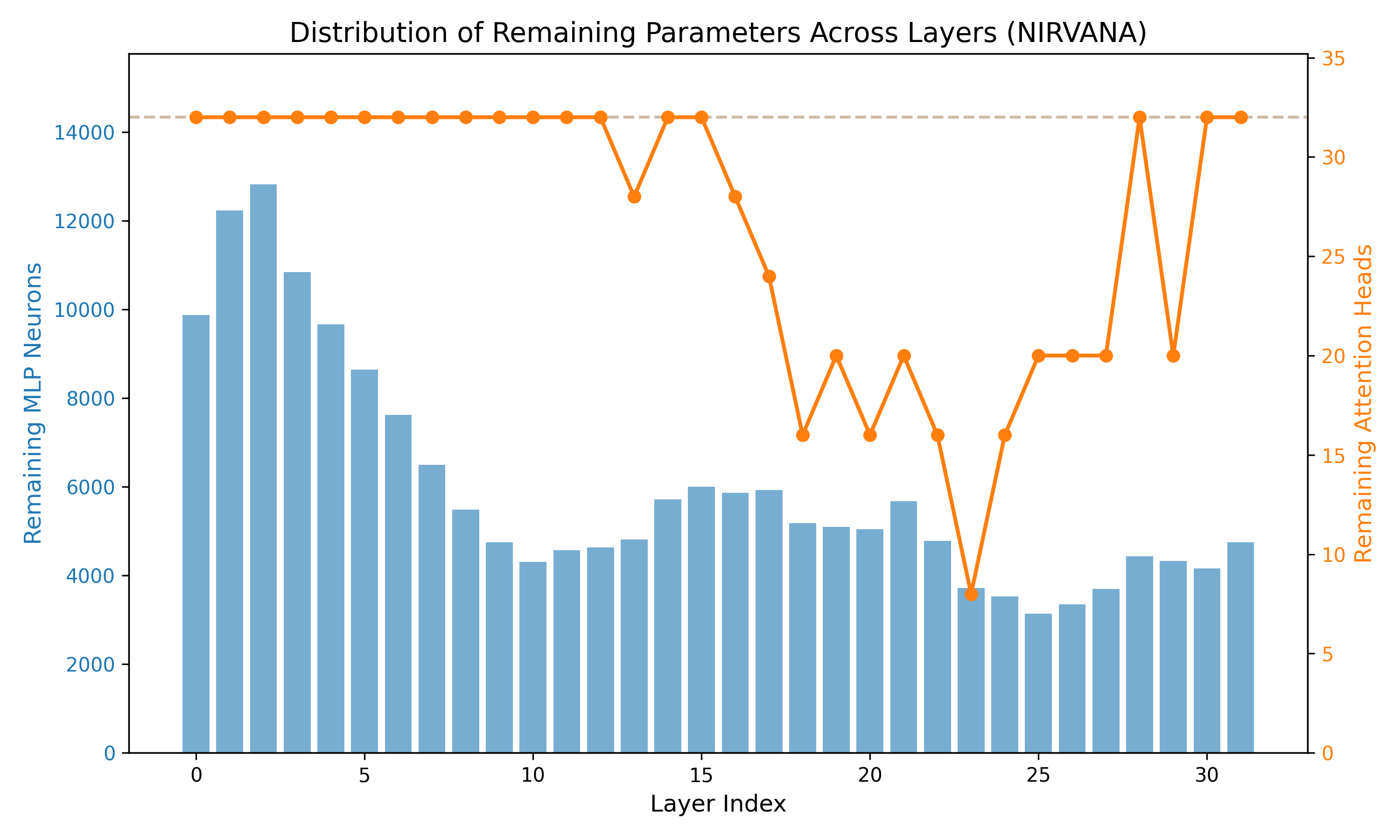}
    \caption{Layer distribution of the pruned Llama3.1-8B at 50\% sparsity levle.}
    \label{fig:layer_dist}
\end{figure}

To visualize the pruning patterns, we provide a layer-wise density plot for the Llama-3.1-8B model at 50\% sparsity in \cref{fig:layer_dist}. The plot reveals a non-uniform strategy where NIRVANA preserves the critical input/output layers while pruning the redundant middle-to-deep layers, which aligns with recent findings \citep{shortgpt}.

\begin{figure}
    \centering
    \includegraphics[width=0.9\linewidth]{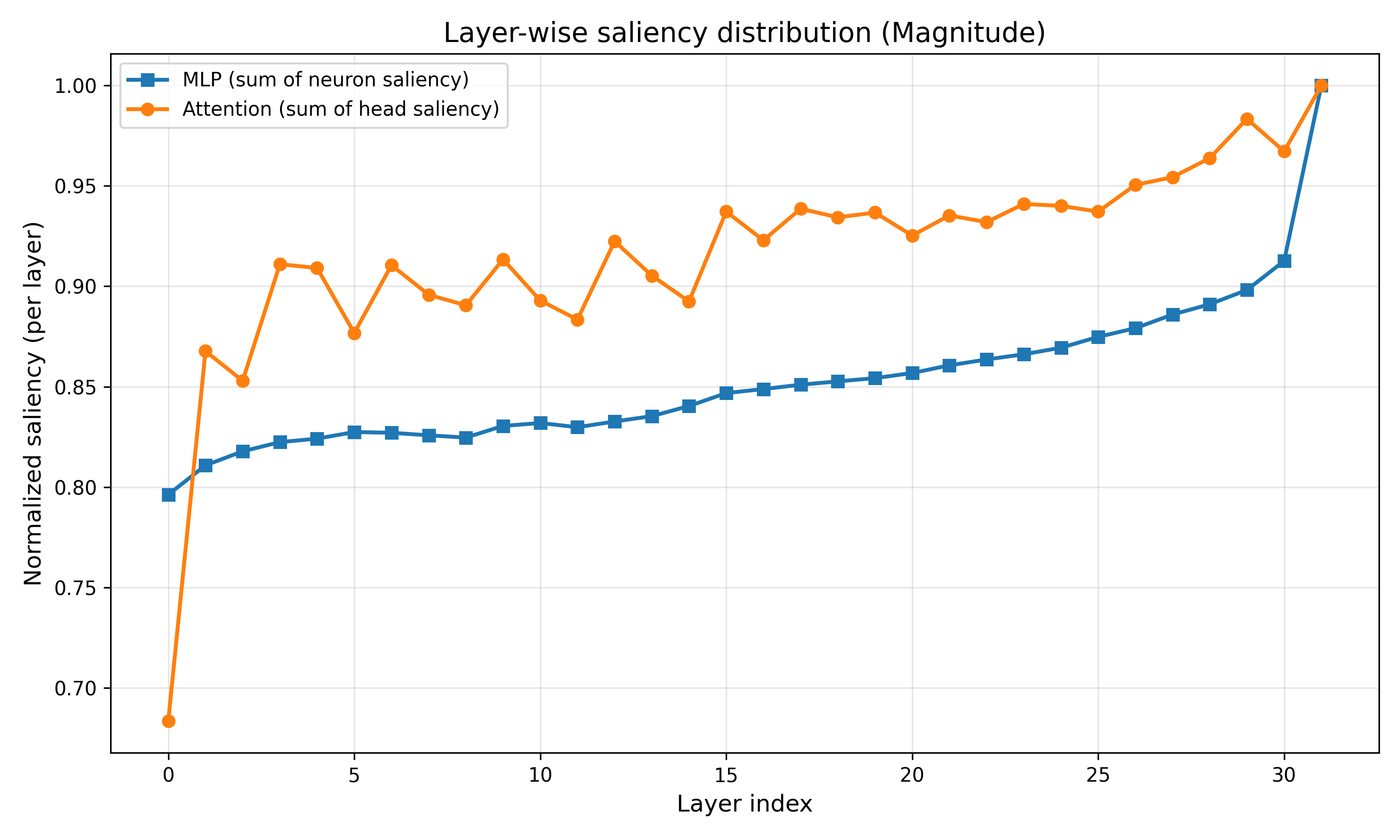}
    \caption{The distribution of the magnitude-based saliency scores across layers on Llama3.1-8B.}
    \label{fig:mag_score}
\end{figure}

\begin{figure}
    \centering
    \includegraphics[width=0.8\linewidth]{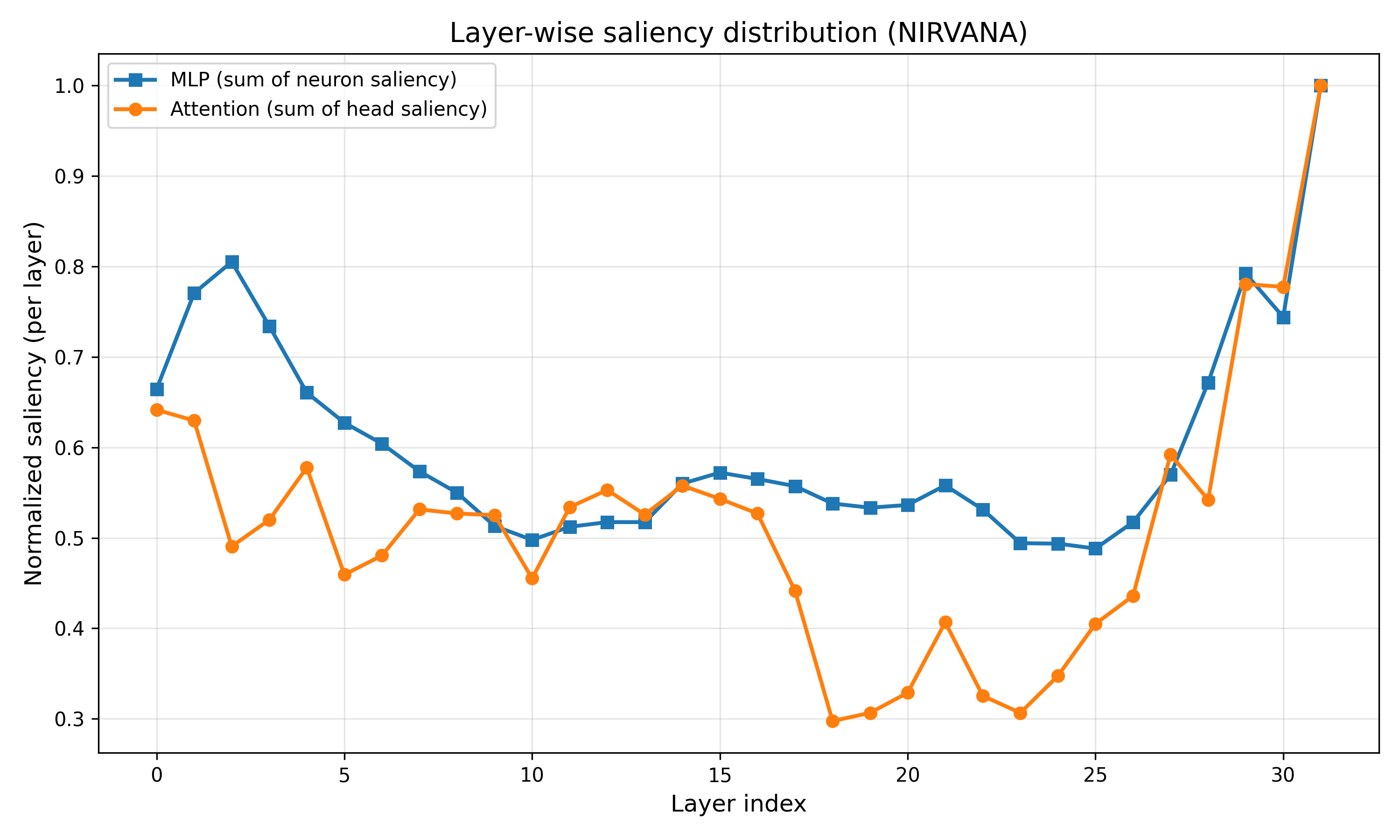}
    \caption{The distribution of NIRVANA's saliency scores across layers on Llama3.1-8B.}
    \label{fig:nir_score}
\end{figure}

As for saliency patterns, the magnitude-based distribution in \cref{fig:mag_score} shows both attention and MLP scores increasing almost monotonically with depth, treating deeper layers as more important. In contrast, the NIRVANA distribution in \cref{fig:nir_score} exhibits a non-monotonic, more flexible depth pattern: it assigns relatively high saliency to the input/output layers while selectively up- or down-weighting different middle blocks, suggesting that the NTK-guided criterion adaptively targets redundant regions.

\newpage

\section{ViT Compatibility}

While our current work focuses on LLMs, NIRVANA's theoretical framework is highly generalizable to non-LM architectures, particularly Vision Transformers (ViTs).

1. The output-gradient based saliency $S(W) = |\frac{\partial f}{\partial W} \cdot W|$, is modality-agnostic. It relies only on the differentiability of the model function $f$ with respect to its weights, regardless of whether the input calibration data consists of tokens or image patches.
2. Since ViTs share the same fundamental Transformer backbone as LLMs, our structured pruning mechanisms are directly transferrable:
   - Saliency Aggregation: The logic for grouping weights into "Heads" or "Neurons" remains identical.
   - Adaptive Allocation ($\gamma$): The analytic derivation for balancing Attention vs. MLP pruning (Appendix F) is based on the architectural definition of the Transformer block, making it applicable to ViT architectures.
   - Optimization Dynamics: ViTs are also commonly trained with Adam/AdamW, preserving the validity of our Adam-NTK stability analysis.

Given this strong mathematical and architectural alignment, extending NIRVANA to vision tasks would follow the established precedent of cross-pollination seen in methods like SNIP \citep{lee2019snipsingleshotnetworkpruning} and NTK-SAP \citep{wang2023ntksapimprovingneuralnetwork}. We view the adaptation of NIRVANA to ViT as a promising and straightforward future direction.

\section{Comparison with Unstructured/Semi-structured Pruning}
\label{app:up_sp}

While Table 7 in the Appendix already compares NIRVANA with Wanda (unstructured) on T5, we provide a more detailed trade-off analysis on Llama-3.1-8B at 50\% sparsity against SparseGPT (unstructured) and MaskLLM (semi-structured 2:4) as follows:

\begin{table}[H]
    \centering
    \resizebox{\linewidth}{!}{
    \begin{tabular}{lcccccc}
    \toprule
      Model   &  Type & WikiT & PTB   & LambD & Pruning Cost                    & Inference Speedup  \\
    \midrule
SparseGPT&	Unstructured	&30.15	&43.87	&52.10&	Moderate (600s)	&None (Irregular) \\
MaskLLM	&Semi-Structured&	12.23&	20.28&	27.78&	Very High (Trains on 2B tokens)&	Restricted (Requires 2:4 HW)\\
NIRVANA	&Structured	&48.94&	70.11	&70.11&	Negligible (< 2s, One-shot)&	High (Guaranteed)\\
\bottomrule
    \end{tabular}
    }
    \caption{Quantitative analysis of the generated examples using ROUGE and BERTScore.}
    \label{tab:pruning_compare}
\end{table}

It is expected that unstructured (SparseGPT) or semi-structured (MaskLLM) methods achieve lower perplexity, as they are less constrained than removing entire structural units. However, NIRVANA targets structured pruning, which is the only paradigm that guarantees latency reduction on general-purpose GPUs without specialized hardware support (as shown in our Table 3 latency results). Additionally, MaskLLM needs to train the model to get their optimal computed task on 2B tokens, with their whole process requiring a huge amount of computational time and memory costs. By contrast, NIRVANA is a one-shot pruning method whose whole process can be efficiently completed within 2 seconds, using 40 GB of memory on a single A100.

\section{The Use of Large Language Models}

In this work, we have used the LLMs to help refine the paper writing and check grammar errors.

\end{document}